\documentclass[preprint,12pt]{elsarticle}

\usepackage{amssymb}

\usepackage{amsmath,amsfonts}
\usepackage{algorithm}
\usepackage{array}
\usepackage[caption=false,font=normalsize,labelfont=sf,textfont=sf]{subfig}
\usepackage{textcomp}
\usepackage{stfloats}
\usepackage{url}
\usepackage{verbatim}
\usepackage{graphicx}
\usepackage{xcolor}
\usepackage{float}

\def\bs{\boldsymbol}

\usepackage{algpseudocode}

\newcommand\Algphase[1]{%
\vspace*{-.7\baselineskip}\Statex\hspace*{\dimexpr-\algorithmicindent-2pt\relax}\rule{\linewidth}{0.4pt}%
\Statex\hspace*{-\algorithmicindent}\textbf{#1}%
\vspace*{-.7\baselineskip}\Statex\hspace*{\dimexpr-\algorithmicindent-2pt\relax}
}

\newcommand{\IndentedAlgorithm}[1]{%
    \hspace{1em} 
    \begin{algorithmic}[1]
    #1
    \end{algorithmic}
    \vspace{-\baselineskip} 
    \hspace{1em} 
}

\usepackage{booktabs} 
\usepackage{comment}
\usepackage{ulem}
\newtheorem{theorem}{Theorem}

\newtheorem{remark}{Remark}

\newtheorem{definition}{Definition}

\usepackage[numbers]{natbib}  
\renewcommand{\thefootnote}{\arabic{footnote}}
\usepackage[colorlinks=true, allcolors=blue]{hyperref}

\usepackage{balance}

\begin{document}

\begin{frontmatter}


\renewcommand{\thefootnote}{\fnsymbol{footnote}}
\title{Beyond Uniform Deletion: A Data Value-Weighted Framework for Certified Machine Unlearning}

\author[label1]{Lisong He}
\affiliation[label1]{organization={Department of Information Systems and Intelligent Business},
            addressline={Xi'an Jiaotong University},
            city={Xi'an},
            postcode={710049},
            state={Shaanxi},
            country={China}}

\author[label2]{Yi Yang\footnote{Corresponding author}}

\affiliation[label2]{organization={Department of Information Systems},
            addressline={Arizona State University},
            city={Tempe},
            postcode={85287},
            state={AZ},
            country={USA}}

\author[label3]{Xiangyu Chang}

\affiliation[label3]{organization={Department of Information Systems and Intelligent Business},
            addressline={Xi'an Jiaotong University},
            city={Xi'an},
            postcode={710049},
            state={Shaanxi},
            country={China}}
            
\begin{abstract}
As the right to be forgotten becomes legislated worldwide, machine unlearning mechanisms have emerged to efficiently update models for data deletion and enhance user privacy protection.
However, existing machine unlearning algorithms frequently neglect the fact that different data points may contribute unequally to model performance (i.e., heterogeneous data values). 
Treat them equally in machine unlearning procedure can potentially degrading the performance of updated models.
To address this limitation, we propose Data Value-Weighted Unlearning (\texttt{DVWU}), a general unlearning framework that accounts for data value heterogeneity into the unlearning process. 
Specifically, we design a weighting strategy based on data values, which are then integrated into the unlearning procedure to enable differentiated unlearning for data points with varying utility to the model.
The \texttt{DVWU} framework can be broadly adapted to various existing machine unlearning methods.
We use the one-step Newton update as an example for implementation, developing both output and objective perturbation algorithms to achieve certified unlearning. 
Experiments on both synthetic and real-world datasets demonstrate that our methods achieve superior predictive performance and robustness compared to conventional unlearning approaches.
We further show the extensibility of our framework on gradient ascent method by incorporating the proposed weighting strategy  into the gradient terms, highlighting the adaptability of \texttt{DVWU} for broader gradient-based deep unlearning methods.

\end{abstract}


\end{frontmatter}

\section{Introduction}

    Many businesses and organizations leverage user data to train machine learning (ML) models across various applications.    
    However, users may request the removal of their data from both the database and the models due to privacy concerns. 
    This raises critical questions: \textit{how can we protect the privacy of users who choose to exit the system}, and more importantly, \textit{how can we effectively update ML models in response to such removal requests?}
    
    In recent years, an increasing number of privacy-related laws and regulations mandating user data withdrawal upon request, known legally as the ``Right to be Forgotten", have been established worldwide.
    Examples include the General Data Protection Regulation (GDPR) \cite{MagorzataMagdziarczyk2019RightTB} in the European Union  and the California Consumer Privacy Act (CCPA) \cite{California} in the United States. 
   
    Importantly, the ``Right to be Forgotten" requires not only the deletion of personal data itself but also the removal of its influence on models. 
   
    Although retraining a model from scratch after each deletion request technically satisfies the influence removal requirement, it is typically computationally expensive and inefficient. Moreover, this approach does not inherently prevent privacy risks: if attackers can access the model both before and after retraining, the deleted user data may still be inferred through reverse inference~\cite{MinChen2021WhenMU}.
    To address these challenges, the ML community has turned to \textit{machine unlearning}, which seeks to efficiently remove the influence of specific data points from trained models~\cite{YinzhiCao2015TowardsMS}.
    
    Machine unlearning aims to eliminate targeted data (influence) from a model without the cost of full retraining, while preserving both utility and privacy. 
    Existing research in this area primarily follows two directions: exact unlearning \cite{LucasBourtoule2019MachineU,AdityaGolatkar2020MixedPrivacyFI}, which guarantees complete removal of a data point's influence through partial retraining, and approximate unlearning \cite{PangWeiKoh2017UnderstandingBP,RyanGiordano2018ASA,ChuanGuo2019CertifiedDR,AyushKTarun2021FastYE}, which offers efficient model updates in an approximate manner with provable error bounds. 
   
    Although existing machine unlearning algorithms enable efficient data removal and model updates, they often lead to degraded overall performance.
    This issue has been demonstrated in several prior works~\cite{YinzhiCao2015TowardsMS,MinChen2021WhenMU,Schelter21He,pawelczyk2022trade,Tian2024DeRDaVa}.
    This degradation may stem from the unequal contribution of individual data points to the model's predictive capability--removing high--impact points can result in substantial increases in the loss function and lead to significant performance declines.
    As a result, the non-differentiated deletion of data--commonly employed in conventional  unlearning methods--can significantly compromise both  accuracy and robustness of the resulting model. 
    We further illustrate this through the following motivating example.

\vspace{2pt}
  
\begin{figure}[!ht]
    \centering
\subfloat[]{\includegraphics[width=0.5\linewidth]{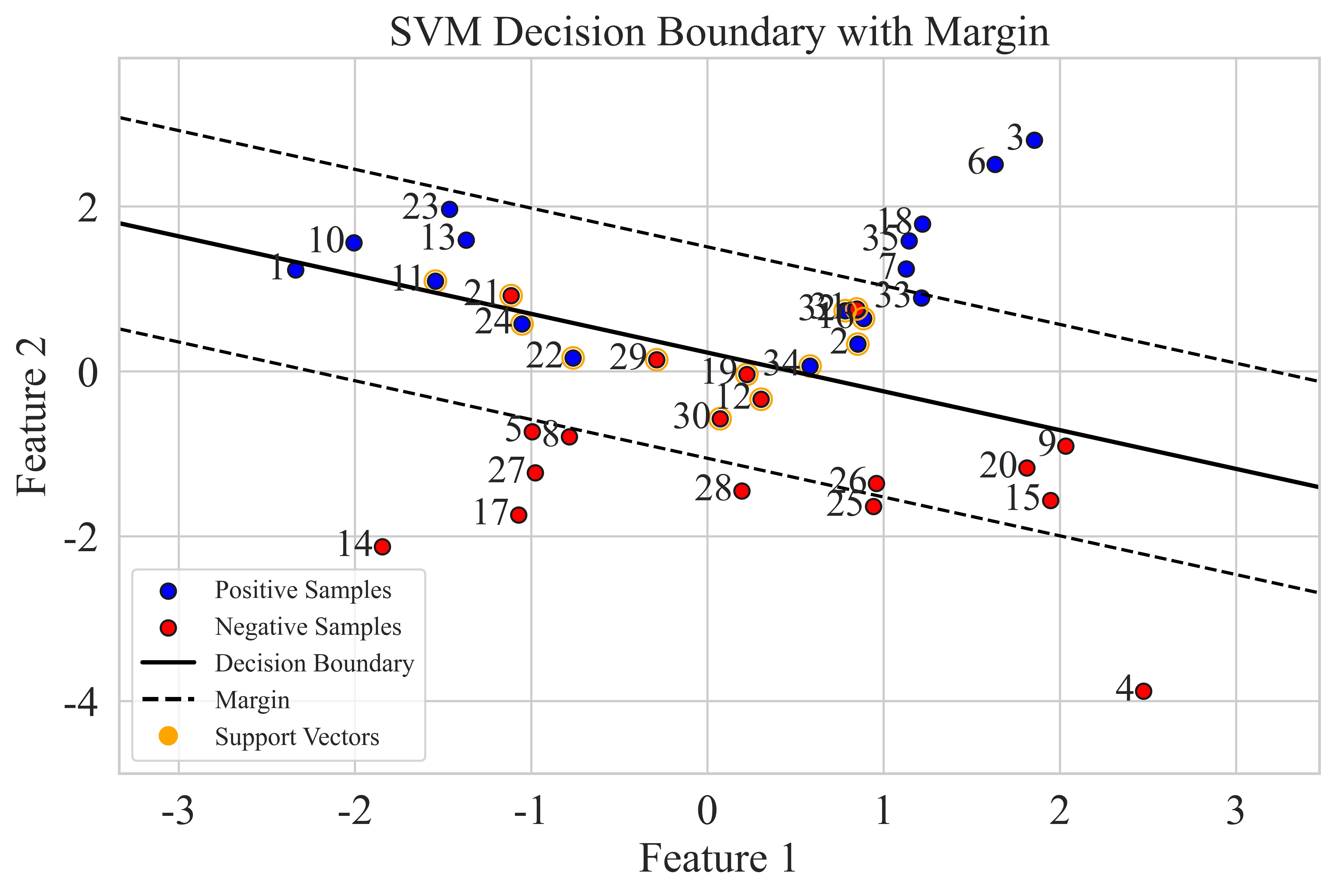}\label{SVM_case}}%
    \hfil
\subfloat[]{\includegraphics[width=0.5\linewidth]{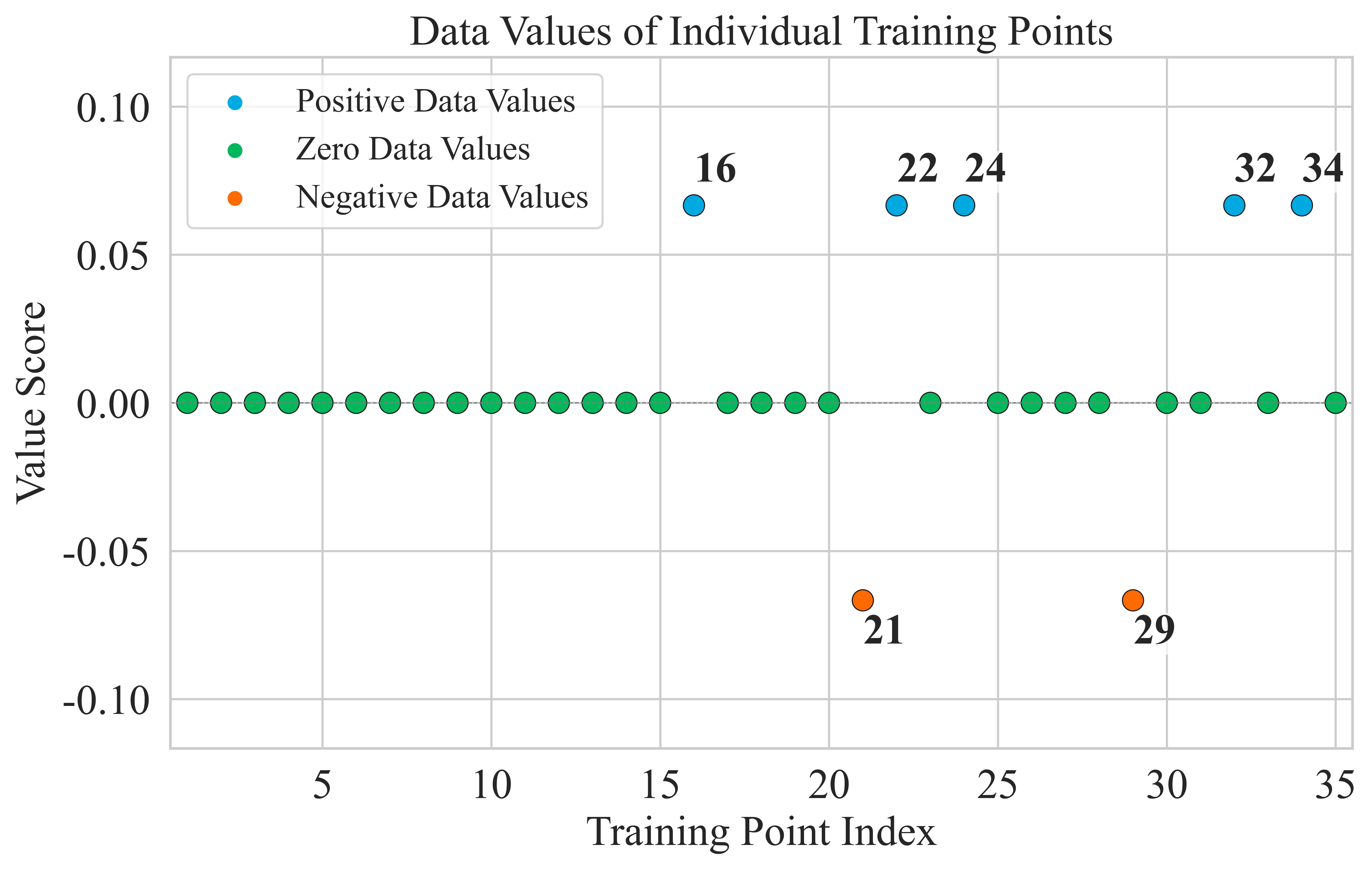}\label{SVM_case2}}%
    \caption{\label{fig:DVExample}  Motivating example: data value in SVM classification.
(a) Decision boundary and margin of a linear SVM trained on 35 synthetic samples.
(b) Leave-one-out data value scores for each training sample, where positive values indicate that removing the point degrades performance, and negative values indicate an improvement.}
    \end{figure}

\noindent \textbf{Motivation Example:} We consider a simple binary classification problem to illustrate how the impact of data removal can vary across instances.\footnote{See Section~\ref{symethod} for dataset and task details.} Figure~\ref{fig:DVExample}\subref{SVM_case} visualizes the decision boundary and margin of a linear SVM trained on 35 samples, along with the leave-one-out~\cite{RDennisCook1977DetectionOI} data value\footnote{See Section~\ref{sec:datavalueassessment} for more details on data value evaluation.} scores for each training point in Figure~\ref{fig:DVExample}\subref{SVM_case2}.
It can be observed that individual data points contribute unequally to model performance--some positively, some negatively, and others not at all.
    As a result, the effect of removing a data point on model performance varies accordingly. For instance, data points indexed as 21 and 29 have negative data values, indicating that their removal would improve model performance. In contrast, points such as 16, 22, 24, 32, and 34 have positive data values, suggesting that removing them would degrade the model's accuracy. 
    Most of the remaining points have zero data values, indicating their negligible influence on the model performance when removal.
    Since a data value quantifies the impact of removing a data point on model performance, and unlearning is precisely the process of eliminating the influence of such points, the heterogeneity of data values naturally explains potential performance fluctuation in unlearning. 
    When the deleted set contains many positively contributing samples (i.e., high-data-value points), unlearning inevitably removes their beneficial effects, thereby degrading the updated model’s performance.
    These observations highlight that, to better preserve model performance after deletions, unlearning strategies should account for each point's data value--an important factor typically overlooked in existing approaches.

    This paper aims to enhance the performance and robustness of machine unlearning by explicitly accounting for the heterogeneous contribution of data points to model performance.
    Specifically, we investigate the following research question:
    
    \textit{How can we design data value-aware machine unlearning methods that mitigate the adverse effects of data deletion on model performance?}
    
\begin{figure*}[!ht]
        \centering \includegraphics[width=\linewidth]{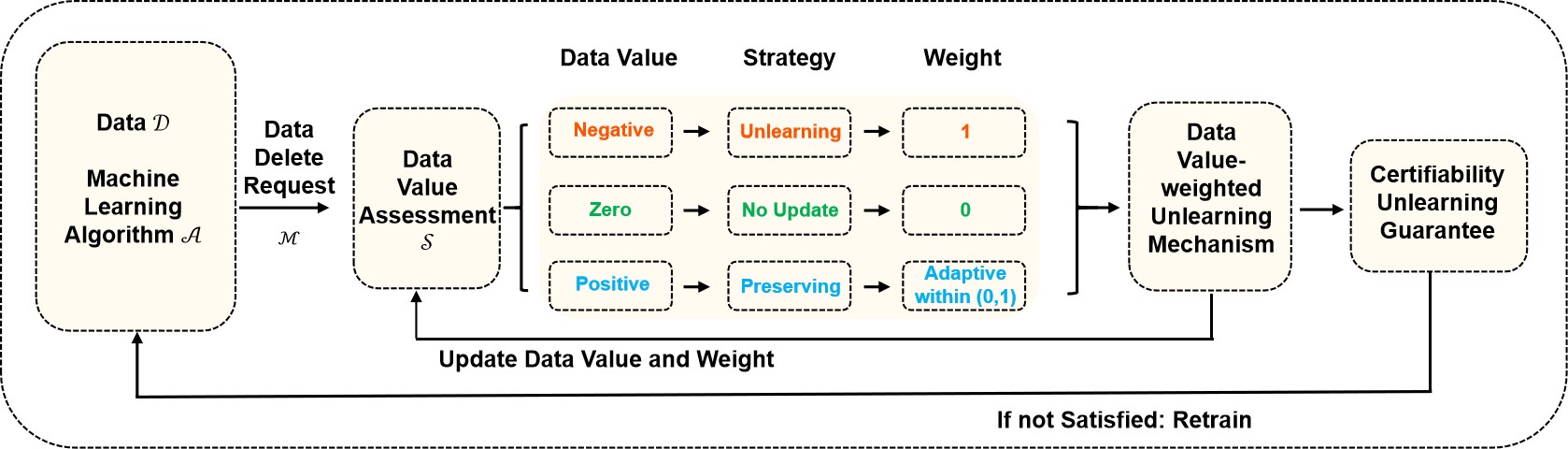}
        \caption{\texttt{DVWU} framework.}
        \label{fig:Framework}
    \end{figure*}

We address this question by incorporating data value into the unlearning process and proposing a general framework called Data Value-Weighted Unlearning (\texttt{DVWU}), as illustrated in Figure \ref{fig:Framework}.

First, given a trained ML model, the data value of each individual data point is estimated using established data valuation methods (e.g.,~\cite{RDennisCook1977DetectionOI,Jia_DaoETSDV2019,zhang2023dynamic}).
Based on these data values, an adaptive weighting strategy is employed: harmful points (i.e., those with negative data value) are assigned a weight of $1$ and subsequently unlearned; valueless points (i.e., with zero data value) are removed by assigning them a weight of $0$; and beneficial points (i.e., those with positive data value) are proportionally downweighted according to their data value.
This weighting strategy is incorporated into the model update process, enabling data value-weighted unlearning when a deletion request for specific data points is received.
After each model update, \texttt{DVWU} performs certified verification to ensure compliance with the deletion request. Subsequently, the data values and corresponding weights of the remaining training samples are recalculated and updated, preparing the model for future deletion requests.

The proposed \texttt{DVWU} framework can be adapted to various existing machine unlearning methods. 
As an example, we use the one-step Newton method~\cite{ChuanGuo2019CertifiedDR}, a classical and widely-used machine unlearning approach to implement \texttt{DVWU}.
We construct  algorithms based on both output perturbation and objective perturbation, providing $(\epsilon, \delta)$-certified unlearning guarantees. 
Finally, we verify the effectiveness and efficiency of our method on both synthetic and real datasets.

    The core contributions of this paper are summarized as follows:
    \begin{itemize}
     \item Motivated by previous observations that direct applications of machine unlearning algorithms can degrade updated model performance, we reveal a connection between this phenomenon and the heterogeneity of data values.     
\item We incorporate data value heterogeneity into the machine unlearning process and propose a novel framework, \texttt{DVWU}. Specifically, we first design a weighting strategy based on data values, which are then integrated into the unlearning procedure to enable differentiated unlearning for data points with varying utility to the model. 
\item We implement \texttt{DVWU} using the one-step Newton update as an example and develop two certified unlearning algorithms based on output perturbation and objective perturbation. 
They provide guarantee on  {certified} removal without significantly sacrificing model performance.

    \item  
    We empirically validate the effectiveness of the proposed framework on both synthetic and real datasets. 
    \texttt{DVWU} outperforms standard unlearning methods and retraining in terms of performance and  robustness, while maintaining strong resilience to noise and data imbalance. Furthermore, we verify the applicability and extensibility of our framework on both linear models and non-linear models, such as deep unlearning.
   
    \end{itemize}

   The remainder of this paper is organized as follows.
    Section~\ref{sec_relatedwork} surveys the related literature.
    Section~\ref{sec_setting} introduces the fundamentals of machine unlearning and presents the one-step Newton update as a representative example of the unlearning mechanism.
    Section~\ref{dvwust} details the proposed weighting strategy and the \texttt{DVWU} framework.
    Section~\ref{alg} implements  \texttt{DVWU}  using the one-step Newton update and presents the corresponding certified unlearning algorithms.
    Section~\ref{sec_experiment} shows the experimental study. Section~\ref{sec_conclusion}  concludes. 
    All technical proofs are provided in the Appendix.

\section{Related Work}\label{sec_relatedwork}
\subsection{Machine Unlearning}\label{sec_MachineUnlearning}
 
    The concept of \textit{machine unlearning} was first introduced by \cite{YinzhiCao2015TowardsMS}. 
    Since then, the field has evolved, with several unlearning methods proposed across various ML paradigms.
    Generally, unlearning algorithms can be categorized into two groups: exact unlearning algorithms and approximate unlearning algorithms~\cite{SalvatoreMercuri2022AnIT}. 

    Exact unlearning algorithms are designed to reduce the cost of model updates following data deletion by structuring the initial training phase.
    For instance, SISA~\cite{LucasBourtoule2019MachineU} partitions the training dataset into multiple subsets, trains an individual base model on each subset, and then ensembles them to make predictions. 
    Upon receiving a deletion request, only the base model associated with the subset containing the targeted data needs to be retrained, significantly lowering the update cost compared to full retraining.
    This approach is later extended to tree-based models, where DaRE~\cite{JonathanBrophy2020MachineUF} is proposed to enable efficient unlearning specifically for decision trees and random forests.

    Approximate unlearning algorithms, on the other hand, manipulate the model parameters learned from the full dataset to approximate those resulting from retraining after removing specific data points.
    This allows for efficient updates without retraining, significantly reducing computational costs.
    For example, \cite{PangWeiKoh2017UnderstandingBP} and \cite{RyanGiordano2018ASA} employ the influence functions to approximate the model parameters that would be obtained by retraining on the dataset with specific data points removed.
    \cite{ChuanGuo2019CertifiedDR} uses the one-step Newton update for parameter approximation. 
    Other works include DeltaGrad \cite{YinjunWu2020DeltaGradRR}, which uses cached gradients to efficiently approximate parameters,  particularly for ML models trained via gradient descent, and the method by Liu \textit{et al.}~\cite{YangLiu2023BackdoorDW}, which adopts gradient ascent for model updates.
    
    To ensure that the results of the unlearning algorithm are similar to those obtained by retraining on the remaining data, the notion of \textit{certified removal}~\cite{ChuanGuo2019CertifiedDR} or \textit{certified unlearning}~\cite{Sekhari2021RememberWY} was proposed, which is inspired by differential privacy~\cite{Dwork2014TheAF}.
    This is commonly achieved by adding perturbations to the loss function~\cite {ChuanGuo2019CertifiedDR,EliChien2022CertifiedGU} or the output~\cite {Sekhari2021RememberWY,YinjunWu2020DeltaGradRR} of the updated model.

    Although existing unlearning methods offer efficient solutions for removing the influence of data from trained models, they may inadvertently degrade model performance—particularly when high informative data points are removed.
    Prior studies show that deleting such critical points can significantly weaken a model's 
    predictive accuracy~\cite{YinzhiCao2015TowardsMS,MinChen2021WhenMU,Schelter21He,pawelczyk2022trade}, and repeated or large-scale deletions can further impair overall model quality~\cite{Tian2024DeRDaVa}.
    This performance degradation introduces a robustness challenge, as models are more sensitive to the loss of informative samples.
    To address this limitation, our method explicitly considers the impact of each data point on model performance.
    By designing and incorporating a data value-based weighting strategy into the unlearning process, we preserve the contribution of valuable data (while still satisfying unlearning requirements), thereby maintaining model robustness and mitigating the adverse effects of data deletion.

\subsection{Data Value Assessment}\label{sec:datavalueassessment}
    A variety of methods have been proposed to assess data value in ML, including influence functions \cite{influenceoo, RDennisCook1977DetectionOI} and the Shapley value \cite{Shapley19977AV}.
    Influence functions \cite{influenceoo, RDennisCook1977DetectionOI} typically measure the change in model performance when individual training points are removed—a process also referred to as the leave-one-out.

    Shapley value-based methods, originally developed in the context of cooperative game theory \cite{Shapley19977AV}, have been adopted for data asset pricing in ML \cite{Jia_DaoETSDV2019, Wang_APATAD_2020}. These methods quantify the marginal contribution of each data point by evaluating its utility across all possible subsets of the training data. 
    However, the exact calculation of Shapley values is computationally expensive, as the number of utility function evaluations grows exponentially with the dataset size \cite{Lundberg_Lee_2017}.
    To address this, several fast approximate estimation methods have been proposed, such as Monte Carlo sampling for approximate Shapley values estimation \cite{Lundberg_Lee_2017, Burgess_Chapman_2021} and KNN-SV \cite{Jia_DaoETSDV2019}, which offer efficient and accurate estimation of Shapley values.
   
    Furthermore, several methods aim to dynamically estimate Shapley values in scenarios where the dataset evolves over time, such as when data is added or deleted. 
    For example, HSV-KNN and HSV-KNN+ \cite{zhang2023dynamic} extend KNN-based Shapley estimation to dynamic settings.
    \cite{xia2023equitable} introduces the Delta-based method and Batched Delta-based method for rapid dynamic approximation of Shapley values.
    Additionally, \cite{Tian2024DeRDaVa} develops an estimation approach that assumes the probability of data deletion can be predicted in advance, allowing the method to account for the expected impact of deletion on Shapley values.

    In this paper, we employ three data value assessment methods—leave-one-out~\cite{RDennisCook1977DetectionOI}, KNN-SV~\cite{Jia_DaoETSDV2019}, and HSV-KNN~\cite{zhang2023dynamic}—to separately measure the impact of each data point on model performance, i.e., its data value.
    Based on data values, we design weighting strategies and incorporate them into the machine unlearning framework.

\section{Problem Setting}\label{sec_setting}

\subsection{Machine Unlearning Definition}

Let $\mathcal{D}=\{\bs{z}_i=(\bs{x}_{i},y_{i})\}_{i=1}^n\subseteq\mathbb{R}^d\times\mathbb{R}$ be a fixed training data set, and $\mathcal{A}$ be a learning algorithm that trains on $\mathcal{D}$ and produces a model $\mathcal{A}(\mathcal{D}) \in \mathcal{H}$, where $\mathcal{H}$ is the hypothesis space of models.
A machine unlearning algorithm aims to remove the influence of a subset of data $\mathcal{M} \subset \mathcal{D}$ from $\mathcal{A}(\mathcal{D})$. 
It can be viewed as an update mechanism $\mathcal{U}$ that updates $\mathcal{A}(\mathcal{D})$ to $\mathcal{U}(\mathcal{A}(\mathcal{D}), \mathcal{M})$. 
Successful removal implies that the output of $\mathcal{U}$ should be difficult to distinguish from the output of $\mathcal{A}$ trained on the reduced data set $\mathcal{D} \backslash \mathcal{M}$ {(i.e., $\mathcal{A}(\mathcal{D} \backslash \mathcal{M})$).

\subsection{One-Step Newton Update}
As discussed in Section \ref{sec_MachineUnlearning}, several model update mechanisms have been developed for machine unlearning, such as SISA \cite{LucasBourtoule2019MachineU}, the one-step Newton update \cite{ChuanGuo2019CertifiedDR}, and the influence method \cite{PangWeiKoh2017UnderstandingBP,RyanGiordano2018ASA}. 
In this section, we use the one-step Newton update as an example to illustrate how model updates are performed by unlearning.

Originally introduced by Beckman \citep{RichardJBeckman1974TheDO} and Cook \citep{RDennisCook1977DetectionOI} in the context of data anomaly measurement, the one-step Newton update evaluates the anomaly of a sample by analyzing the changes in model parameters after its removal.
Over time, it has been adapted to efficiently estimate model updates in machine unlearning \citep{ChuanGuo2019CertifiedDR,Sekhari2021RememberWY}.

For illustrative purposes, consider a learning algorithm $\mathcal{A}$ that aims to minimize the regularized empirical risk:
    \begin{equation}
     L(\boldsymbol{w}; \mathcal{D}) = \frac{1}{n} \sum_{i=1}^n \ell(\boldsymbol{w},
     \bs{z}_i) + \frac{\lambda}{2} \|\boldsymbol{w}\|^2,
     \label{eq:lossf}
     \end{equation}
    where $\boldsymbol{w}$ is the model parameters, $ \ell(\cdot)$ is a convex loss that is differentiable everywhere, and $\lambda>0$ is the regularization parameter.
    
Let $\boldsymbol{w}^* =\mathcal{A}(\mathcal{D})=\arg \min_{\boldsymbol{w}} L(\boldsymbol{w}; \mathcal{D})$ be the optimal solution and $\mathcal{M} \subset \mathcal{D}$ be the data points to be deleted with $\mathcal{M}=\{\bs{z}_i\}_{i=1}^m$\footnote{Without loss of generality, we assume  that we aim to remove the first $m$ training samples.}.
Then, the approximated model parameters after deleting $\mathcal{M}$ by one-step Newton update is as follows \cite{Sekhari2021RememberWY}:
    \begin{equation}
    \boldsymbol{w}^{\prime} = \boldsymbol{w}^* + \frac{m}{n-m}[\boldsymbol{H}_{\boldsymbol{w}^*}^{\mathcal{D}\setminus\mathcal{M}}]^{-1} \nabla L(\boldsymbol{w}^*; \mathcal{M}).
    \label{eq:newtonM}
    \end{equation}
    Here, $\boldsymbol{H}_{\boldsymbol{w}^*}^{\mathcal{D}\setminus\mathcal{M}}=\nabla^2 L(\boldsymbol{w}^*; \mathcal{D}\setminus\mathcal{M})$ represents the Hessian matrix of $ L(\cdot; \mathcal{D}\setminus\mathcal{M})$ at $\boldsymbol{w}^*$. 
    $\nabla L(\boldsymbol{w}^*; \mathcal{M})$ denotes the gradient of $L$ with respect to the data being removed, evaluated at $\boldsymbol{w}^*$.
    
    For a trained ML model $\mathcal{A}(\mathcal{D})$, the model parameters $\boldsymbol{w}^*$ and the Hessian of $L(\cdot; \mathcal{D})$ at $\boldsymbol{w}^*$ (i.e., $\boldsymbol{H}_{\boldsymbol{w}^*}^{\mathcal{D}}$), are saved.  
    When a request to delete data $\mathcal{M}$ is received, $\nabla L(\boldsymbol{w}^*; \mathcal{M})$ and $\boldsymbol{H}_{\boldsymbol{w}^*}^{\mathcal{D}\setminus\mathcal{M}}$ are calculated based on $\boldsymbol{w}^*$ and $\boldsymbol{H}_{\boldsymbol{w}^*}^{\mathcal{D}}$.  
    The updated model parameters, after data deletion, are then approximated by $\boldsymbol{w}^{\prime}$ using \eqref{eq:newtonM}.
    This approach enables model updates through machine unlearning without requiring full retraining.
    \cite{ChuanGuo2019CertifiedDR,Koh2019OnTA} demonstrate that the one-step Newton update is significantly more time-efficient than retraining while maintaining comparable model performance.

\subsection{Certified Unlearning}

To ensure close alignment between unlearning mechanisms and retraining, the notion of \textit{certified removal}~\cite{ChuanGuo2019CertifiedDR} or \textit{certified unlearning}~\cite{Sekhari2021RememberWY} was proposed.

\begin{definition}\label{Definition_1} Given $\epsilon, \delta\geq 0$, an unlearning mechanism $\mathcal{U}$ performs $(\epsilon,\delta)$-certified unlearning for a learning algorithm $\mathcal{A}$, if 
 for any $\mathcal{T} \subseteq \mathcal{H}$ and $\mathcal{M}\subset \mathcal{D}$:
  \begin{align*}
&\operatorname{Pr}\left[\mathcal{U}\left(\mathcal{A}(\mathcal{D}), \mathcal{M}\right) \in \mathcal{T}\right] \leq e^\epsilon \operatorname{Pr}\left[\mathcal{A}\left(\mathcal{D} \backslash \mathcal{M}\right)\in \mathcal{T}\right]+\delta,\\
&\operatorname{Pr}\left[\mathcal{A}\left(\mathcal{D} \backslash \mathcal{M}\right)\in \mathcal{T}\right] \leq e^\epsilon \operatorname{Pr}\left[\mathcal{U}\left(\mathcal{A}(\mathcal{D}), \mathcal{M}\right) \in \mathcal{T}\right]+\delta.
    \label{eq:CertifiedUl}
    \end{align*}    
\end{definition}

If $\delta = 0$, the unlearning mechanism $\mathcal{U}$ satisfies $\epsilon$-certified unlearning without relaxation.
In contrast, $(\epsilon, \delta)$-certified unlearning allows for a relaxation, allowing the strict $\epsilon$-certified unlearning to be violated with a small probability $\delta$ for certain low-probability events.
 
To achieve certified unlearning,~\cite{ChuanGuo2019CertifiedDR} introduces noise perturbation into the loss function, where the perturbation distribution is determined by the upper bound of the gradient residual norm $ \left\|\nabla L\left(\boldsymbol{w}^{\prime}; \mathcal{D} \setminus \mathcal{M}\right)\right\|_2$. 
In addition, \cite{Sekhari2021RememberWY} adds noise directly to the updated model parameters. 
These approaches are conceptually related to noise injection in differential privacy and can prevent the inference of deleted data~\cite{MinChen2021WhenMU, ChuanGuo2019CertifiedDR}.

\section{Data Value-Weighted Unlearning Framework}\label{dvwust}

    As illustrated in the motivation example (see Figure~\ref{fig:DVExample}), data points contribute unequally to model performance.
    Ignoring these differences and applying machine unlearning uniformly may lead to performance degradation.
    To address this issue, we propose incorporating data value as a weighting factor into the machine unlearning process.
    This section introduces the proposed data value-based weighting strategy and presents the corresponding \texttt{DVWU} framework.

    In a more general setting, we consider a continuous deletion scenario. 
    Let $t \in \{i \in \mathbb{N}^+ \mid i \le T\}$ denote the $t$-th round of data deletion.
    We define:
\begin{itemize}

    \item \textbf{Initial dataset}: Let $\mathcal{D}^0 = \mathcal{D}$ denote the original dataset prior to any deletion.
   
    \item \textbf{Deleted data in round $t$}: 
    Let $\mathcal{M}^t \subseteq \mathcal{D}^{t-1}$ denote the subset of data points removed at the $t$-th deletion round, where the $i$-th deleted point is denoted by $\bs{z}_i^t$.  
    To simplify the presentation, we assume that each deletion round removes the first $m$ data points (i.e., $|\mathcal{M}^t|=m$).\footnote{ This assumption can be easily relaxed. An extension to non-uniform deletion sizes under continuous deletion scenarios is  presented in Appendix A.3.}

    \item \textbf{Remaining dataset after $t$-th deletion}: 
    Let $
    \mathcal{D}^t = \mathcal{D}^{t-1} \setminus \mathcal{M}^t$ denote the remaining dataset after the $t$-th deletion round, where $n^t = |\mathcal{D}^t|$ represents its sample size.

 \item \textbf{Data value}: Let $\boldsymbol{q}^t = \{q_i^t\}_{i=1}^{n^{t-1}}$ denote the set of data values for the dataset $\mathcal{D}^{t-1}$, where $q_i^t$ represents the date value of the $i$-th data point. These values are computed using a data value assessment method $\mathcal{S}$. 

   \item \textbf{Data value weight}: Let $\boldsymbol{v}^t = \{v_i^t\}_{i=1}^{n^{t-1}}$ denote the set of data value weights for the data points in $\mathcal{D}^{t-1}$.
   These weights are derived from $\boldsymbol{q}^t$.
   
\end{itemize}
    
For the $t$-th round deletion, We consider the following optimization problem for model updating, which incorporates data value information:
    \begin{equation}\label{dvwumodel}
        \min_{\boldsymbol{w}}  L(\boldsymbol{w};\mathcal{D}^{t-1})-\sum_{i=1}^{m} v_i^t L\left(\boldsymbol{w} ; z_i^t\right),
    \end{equation}
    where $L$ is defined as in \eqref{eq:lossf}.
    Each weight $v_i^t \in [0, 1]$ reflects the data value of the corresponding point $z_i^t$. 
    By introducing these weights, the unlearning process accounts for the heterogeneity in data value, enabling a more nuanced model update. 
    When $v_i^t = 1$ for all $i$, the formulation reduces to the conventional unlearning setting without data value differentiation. 
    Notably, the optimization problem \eqref{dvwumodel} can be approximately solved by integrating data value weights into existing machine unlearning paradigms.

\subsection{Data Value-Weighting Strategy}\label{sec:weightingstrategy}

    Data value-weighting plays an important role in our framework. In this subsection, we describe how to calculate weights $\boldsymbol{v}^t$ from data values $ \boldsymbol{q}^t$.
    We begin by defining the minimum positive data value in the original dataset as  $q_{\min}^{+} = \min_{i \, : \, q_i^1 > 0} q_i^1$. 
    Then, the weight vector $\boldsymbol{v}^t$ is determined from $\boldsymbol{q}^t$ according to the following strategy. 
    
    For any point $i$:
\begin{itemize}
    \item \textbf{If $q_i^t$ is negative}, its removal leads to performance improvement.
    Thus, its weight $v_i^t$ is set to $1$, indicating that the data point should be directly removed from the training set, and the model is updated using the existing unlearning algorithm.

    \item \textbf{If $q_i^t$ is zero}, its removal has no impact on model performance.    
    Thus, we set $v_i^t=0$, and we simply remove the data from the training set without updating the model.\footnote{Note that in continuous deletion scenarios, if the one-step Newton update is used for unlearning, even though the model parameter does not require updating, the Hessian matrix still needs to be updated by removing the contribution of the point $i$ to ensure the consistency for subsequent unlearning steps.}

    \item \textbf{If $q_i^t$ is positive}, removing this point degrades model performance. 
    Its weight is set to $v_i^t=\frac{\alpha q^{+}_{\min}}{q_i^t}$, where $\alpha \in (0,1]$ is an adjustment coefficient to avoid assigning weight 1 to the data point with the smallest positive data value (i.e., avoiding its direct deletion).
    This strategy ensures that data points with higher positive value receive smaller weights, thereby preserving their contribution to model accuracy and enhancing robustness.
\end{itemize}

    Based on this strategy, the data value weight $v_i^t $ to be used in the $t$-th round deletion  can be determined using the following weighting function $f$:
   
  \begin{equation}
    v_i^t = f(q_i^t) = 
    \begin{cases}
    1, & \text{if } q_i^t < 0,\\
    0, & \text{if } q_i^t = 0, \\
    \dfrac{\alpha q^{+}_{\min}}{q_i^t}, & \text{if } q_i^t > 0.
    \end{cases}
    \label{eq:weightfc}
    \end{equation}
For the $t$-th round deletion, given data $\mathcal{D}^{t-1}$, we can calculate the data value vector $\boldsymbol{q}^t$ using any data value assessment method, and derive the weight vector $\boldsymbol{v}^t$ according to \eqref{eq:weightfc}.
In this way, the data value information associated with the deletion set 
$\mathcal{M}^t$ can be integrated into the model updating process.

\begin{remark}\label{remark1}
The data value vector \( \boldsymbol{q}^t \), and thus the weight function \( \boldsymbol{v}_i^t \) defined in~\eqref{eq:weightfc}, can be determined in either a \textit{static} or \textit{dynamic} manner.
\textit{Static} valuation computes the data value vector once at the initial stage (i.e., \( \boldsymbol{q}^1 \)) and keeps it fixed throughout all following deletion rounds. 
This approach reduces computational cost.
Typical methods such as KNN-SV~\cite{Jia_DaoETSDV2019} and the leave-one-out method~\cite{RDennisCook1977DetectionOI} can be used to estimate static data values.
However, since these values do not reflect changes in the dataset after deletions, they may become less accurate over time.
\textit{Dynamic} valuation, on the other hand, recalculates the data value vector \( \boldsymbol{q}^t \) at each round \( t \), thereby capturing the evolving contribution of each point as the dataset changes. This improves estimation accuracy but incurs a higher computational cost. 
Thus, efficient data value approximation techniques such as HSV-KNN~\cite{zhang2023dynamic}, Delta-based, and Batched Delta-based methods~\cite{xia2023equitable,Tian2024DeRDaVa} can be employed to improve computational efficiency.

\end{remark}

\subsection{Data Value-Weighted Unlearning Framework}

The proposed \texttt{DVWU} framework is outlined in Algorithm~\ref{alg:DVMUFramwork}, which consists of three main components:

\begin{algorithm}[H]
\caption{Data Value-Weighted Unlearning Framework }\label{alg:DVMUFramwork}
\begin{algorithmic}[1]
\Require  Dataset $\mathcal{D}^0$, learning algorithm $\mathcal{A}$, unlearning mechanism $\mathcal{U}$, threshold for certified unlearning verification, the total number of deletion rounds $T$, data value assessment method $\mathcal{S}$, and $\mathcal{M}^1, \mathcal{M}^2, \ldots, \mathcal{M}^T$.
\IndentedAlgorithm{
\Algphase{Model Training \& Data Value Weight Calculation}
\State  Train initial model: $\boldsymbol{w}^0 \gets \mathcal{A}(\mathcal{D}^0)$.
\State Calculate data values: $\boldsymbol{q}^1 \gets \mathcal{S}(\mathcal{D}^0)$.
\State Calculate weights: $\boldsymbol{v}^1 \gets f(\boldsymbol{q}^1)$ using ~\eqref{eq:weightfc}.
\State Save $\boldsymbol{w}^0$ and other information required by $\mathcal{U}$.

\Algphase{Data Value-Weighted Unlearning Mechanism}
\For{$t = 1, 2, \ldots, T$}
    \State Receive deletion request: $\mathcal{M}^t \subset \mathcal{D}^{t-1}$.
    \State Weighted unlearning: $\boldsymbol{w}^t \gets \mathcal{U}(\boldsymbol{w}^{t-1},\mathcal{M}^t, \boldsymbol{v}^t)$.
    
    \State Update dataset: $\mathcal{D}^t \gets \mathcal{D}^{t-1} \setminus \mathcal{M}^t$.  
    \State Recalculate data values: $\boldsymbol{q}^{t+1} \gets \mathcal{S}(\mathcal{D}^{t})$.
    \State Recalculate weights: $\boldsymbol{v}^{t+1} \gets f(\boldsymbol{q}^{t+1})$.

    \Algphase{Certified Unlearning Guarantee}
    \State Compute certification metric (e.g., gradient residual).
    \If{certification metric $\leq \text{threshold}$}
        \State Proceed to next deletion round.
    \Else 
        \State Retrain model: ${\boldsymbol{w}^*}^t \gets \mathcal{A}(\mathcal{D}^t)$.
        \State  For the next iteration: ${\boldsymbol{w}}^t \gets {\boldsymbol{w}^*}^t$.
        \State Reinitialize: $\boldsymbol{q}^{t+1}, \boldsymbol{v}^{t+1}$.
    \EndIf
\EndFor
}
\end{algorithmic}
\end{algorithm}

\begin{itemize}

\item[(1)]\textit{Model Training and Data Value Weight Calculation}: 
    An ML model is first trained on the original dataset, and the information required by the unlearning algorithm (e.g., model parameters, Hessian matrix) is stored.   
    Subsequently, data values are estimated using established methods such as the leave-one-out approach~\cite{RDennisCook1977DetectionOI} and the Shapley value~\cite{Shapley19977AV, Jia_DaoETSDV2019, Wang_APATAD_2020}.  
    Once the data values are obtained, the proposed data value-weighting strategy is applied to convert them into corresponding weights.

    \item[(2)] \textit{Data Value-Weighted Unlearning Mechanism:}
    With the calculated weights and stored model information, we incorporate data value weights into any existing machine unlearning algorithm to guide the model update process. 
    After each update, the data values are recalculated \footnote{{Data values can be recomputed exactly using $\mathcal{S}(\mathcal{D}^t)$, or approximated from the previous value scores $\boldsymbol{q}^{t}$, together with the previous data set $\mathcal{D}^{t-1}$ and the deletion set $\mathcal{M}^t$.}}.
    We will illustrate this step in detail in the next section, using the one-step Newton update as an example.

    \item[(3)] \textit{Certified Unlearning Guarantee:} We ensure certified guarantees via perturbations to model parameters (output perturbation) or the objective function (objective perturbation), following \cite{ChuanGuo2019CertifiedDR,Sekhari2021RememberWY}.
    During the unlearning process, we continuously monitor a verification metric (e.g., gradient residual) against a given  threshold. If the condition is met, unlearning proceeds; otherwise, the model falls back to retraining to maintain the certifiability guarantee.
    
\end{itemize} 
   
 The proposed framework is broadly applicable and can be flexibly integrated into a wide range of existing machine unlearning methods (see  Remark \ref{remark:other_unlearning}) by incorporating data value weights into their update procedures. 
 This enables the resulting unlearning process to better account for heterogeneous data contribution, enhancing both performance and robustness.

\section{\texttt{DVWU} for One-Step Newton Update} \label{alg}

In this section, we use the one-step Newton update as a representative example to implement the \texttt{DVWU} framework.
We begin by introducing a value-weighted update scheme with theoretical guarantees. 
Based on this, we then develop two certified unlearning algorithms, using output perturbation and objective perturbation, respectively.

\subsection{Data Value-Weighted One-Step Newton Update}

We illustrate how \texttt{DVWU} can be applied to the one-step Newton update in the context of continuous data deletions.
In this approach, we assign data value-based weights to the gradients of the data points being deleted, guided by two key considerations.
First, from a theoretical standpoint, the effect of a deleted data point on the model parameters is primarily governed by its first-order gradient in the update formula. Thus, weighting the gradient provides a direct and effective means to control each point's impact on the model update according to its data value.
Second, from a practical perspective, gradient weighting is both computationally efficient and highly extensible. It can be readily integrated into other gradient-based machine unlearning algorithms, making it a general and scalable strategy for incorporating data valuation into unlearning.

    At the deletion round $t$, the weighted gradient for the deletion set $\mathcal{M}^t$  at model parameters $\boldsymbol{w}^{t-1}$ is defined as:

    \begin{equation}
    \begin{aligned}  
    \nabla L_{\boldsymbol{v}}(\boldsymbol{w}^{t-1}; \mathcal{M}^t)  =\frac{1}{m} \sum_{i=1}^m v_i^t\left(\nabla\ell\left(\boldsymbol{w}^{t-1},z_i^t\right)+\lambda \boldsymbol{w}^{t-1}\right),
    \label{eq:dvgradient}    
   \end{aligned}
   \end{equation}
    where $v_i^t$ are the data value weights corresponding to $z_i^t\in\mathcal{M}^t$. 
    
   Based on the weighted gradient, the model parameters are then updated via the following data value-weighted one-step Newton formula: 

   \begin{equation}
    \boldsymbol{w}^t = \boldsymbol{w}^{t-1} +\frac{m}{n-tm} [\boldsymbol{H}_{\boldsymbol{w}^{t-1}}^{\mathcal{D}^t}]^{-1} \nabla L_{\boldsymbol{v}}(\boldsymbol{w}^{t-1}; \mathcal{M}^t),
    \label{eq:newtonMwc}
    \end{equation}
    where $\boldsymbol{H}_{\boldsymbol{w}^{t-1}}^{\mathcal{D}^t}=\nabla^2 L(\boldsymbol{w}^{t-1}; \mathcal{D}^t)$ represents the Hessian matrix of $ L(\cdot; \mathcal{D}^t)$ at $\boldsymbol{w}^{t-1}$. Specifically, $\bs{w}^0=\bs{w}^*$ is the initial model parameter trained on $\mathcal{D}^0$.

    This weighting scheme allows the influence of each data point on $\boldsymbol{w}^t$ to be adjusted according to its data value.  
    If a data point has zero data value, its weight is set to $v_i^t = 0$, resulting in no contribution to the weighted gradient $\nabla L_{\boldsymbol{v}}(\boldsymbol{w}^{t-1}; \mathcal{M}^t)$, and thus no unlearning procedure is required.
    If a data point has a negative data value, it is assigned a weight $v_i^t = 1$ and fully accounted in the gradient, ensuring its influence is removed.
    This helps effectively eliminate the impact of data points considered detrimental to the model. 
    For data points with positive data values, their weights are set inversely proportional to their data value, i.e., a higher data value leads to a smaller weight. 
    Consequently, their influence in the weighted gradient for model update is suppressed, allowing their positive effect on model performance to be largely preserved.

    In what follows, we derive theoretical upper bounds for the gradient residual and parameter gap in continuous data deletions.
    These bounds will support the design of certified unlearning in our proposed \texttt{DVWU} algorithms, which we will discuss later.
    Detailed proofs of Theorems~\ref{theorem_1} and~\ref{theorem_2} are provided in Appendices A.1 and A.2.

\begin{theorem}\label{theorem_1}
Assume that for all $i$, $\|\boldsymbol{x}_i\|_2 \leq 1$. 
Additionally, for all $\boldsymbol{z}_i \in \mathcal{D}$ and $\boldsymbol{w} \in \mathbb{R}^d$, we suppose that the gradient of the individual loss function with respect to $\boldsymbol{w}$ is bounded, i.e., $\|\nabla \ell(\boldsymbol{w}, \boldsymbol{z}_i)\|_2 \leq C$ for some constant $C > 0$.
Furthermore, the Hessian of the individual loss function $\nabla^2 \ell(\boldsymbol{w}, \boldsymbol{z}_i)$ is $\beta$-Lipschitz continuous.
Let $\boldsymbol{w}^t$ denote the model parameters at the $t$-th deletion round, updated using the data value-weighted one-step Newton update. 
Then, we have:
\begin{equation} \left\|\nabla L\left(\boldsymbol{w}^t; \mathcal{D}^t \right)\right\|_2 \leq \frac{4 \beta C^2 m^2 t}{\lambda^2(n-tm)^2}+\frac{4 C m t}{n-tm}.\end{equation}
\end{theorem}

\begin{theorem}\label{theorem_2}
Let $\boldsymbol{w}^t$ denote the model parameters at the $t$-th deletion round, updated using the data value-weighted one-step Newton update, and let ${\boldsymbol{w}^*}^t = \arg \min_{\boldsymbol{w}} L(\boldsymbol{w}; \mathcal{D}^t)$ denote the parameters obtained by retraining on the updated dataset $\mathcal{D}^t$.
Under the same assumptions as in Theorem~\ref{theorem_1}, we have:
\begin{equation}
    \left\|{\boldsymbol{w}^*}^t - \boldsymbol{w}^t\right\|_2 \leq \frac{4 \beta C^2 m^2 t}{\lambda^3(n -tm)^2} + \frac{4 C m t}{\lambda(n - tm)}.
    \label{eq:paramterboundt}
\end{equation}
\end{theorem}

\begin{remark}\label{remark:other_unlearning}
The proposed data value-weighted framework is broadly applicable and can be extended to various unlearning paradigms beyond the one-step Newton update.
For approximate unlearning algorithms--such as the influence function method \citep{PangWeiKoh2017UnderstandingBP,RyanGiordano2018ASA} and gradient ascent \citep{thudi2022unrolling,YangLiu2023BackdoorDW,yao2025large}--the same data value weighting strategy can be applied to the gradient terms in their model update rules.
For instance, in gradient ascent--a method commonly used in deep unlearning--the data value weights can be directly incorporated into the gradient updates, as shown in ~\eqref{eq:dvgradient} (see Appendix B.9 for more details).
For exact unlearning algorithms, such as SISA \citep{LucasBourtoule2019MachineU}, data value weights can be introduced by solving the weighted optimization problem in~\eqref{dvwumodel} within the pre-partitioned subset that contains the deleted samples $\mathcal{M}^t$.
We leave these extensions to interested readers.
\end{remark}

\subsection{\texttt{DVWU} with Output Perturbation}

Certified unlearning can be achieved by applying \texttt{DVWU} with output perturbation, as detailed in Algorithm~\ref{alg:DVMU1}.
First, an ML model is trained by optimizing $L(\boldsymbol{w}; \mathcal{D})$ defined in \eqref{eq:lossf}. The optimal model parameter $\boldsymbol{w}^*$ is saved as $\boldsymbol{w}^0$, and the Hessian of $L$ at $\boldsymbol{w}^*$ is saved as $\boldsymbol{H}^0$. Then, the data value vector $\boldsymbol{q}^1$ is calculated using a data value assessment method $\mathcal{S}$, and the weight vector $\boldsymbol{v}^1$ is derived based on our  weighting strategy introduced in Section~\ref{sec:weightingstrategy}.

In the \texttt{DVWU} mechanism part, the data value-weighted one-step Newton update is applied to approximate the updated model parameters at each round. To improve computational efficiency, the gradient of $L$ with respect to the data $z_i^t$ is calculated only for points whose $v_i^t \neq 0$, as shown in line \ref{deltac}. 
After each update, the data value of the remaining data is recalculated if we use a dynamic data value weight (as discussed in Remark \ref{remark1}).

In the certified unlearning part, noise $\bs{b}^t$ is then added to produce the final model ${\boldsymbol{w}^{\prime}}^t$. It is noteworthy that Algorithm \ref{alg:DVMU1} outputs both the original updated model parameters and the perturbed ones. 
The former can be used for prediction and model iteration, while the latter ensures data privacy when the algorithm is made public. 
To verify that certified unlearning is satisfied, the gradient residual norms $\boldsymbol{g}^{t}$ after each model update are calculated and compared with a given threshold. Since the algorithm guarantees strong certifiability (see the certified unlearning experiment in Section~\ref{uncertifia}), it is not necessary to compute the gradient residual for every deletion, which can save computational costs. 
If the values are below the threshold, unlearning proceeds; otherwise, the model falls back to retraining to preserve the certifiability guarantee. After retraining, the parameter ${\boldsymbol{w}^{*}}^t$ is taken as the exact value of ${\boldsymbol{w}}^t$ and used in subsequent iterations.

Theorem \ref{theorem_cop} establishes the certified unlearning guarantees for Algorithm \ref{alg:DVMU1} (see proof in Appendix A.4).
  
 \begin{theorem}\label{theorem_cop}
    \textit{ Let $\mathcal{A}$ be the learning algorithm that returns ${\boldsymbol{w}^*}^t$, where ${\boldsymbol{w}^*}^t = \arg \min_{\boldsymbol{w}}  L(\boldsymbol{w}; \mathcal{D}^t)$.
    Then, Algorithm \ref{alg:DVMU1}  provides $(\epsilon, \delta)$-certified unlearning of $\mathcal{A}$.}
  \end{theorem}

\begin{algorithm}[!ht]
\caption{\texttt{DVWU} with Output Perturbation in Continuous Deletion}\label{alg:DVMU1}
\begin{algorithmic}[1]
\Require  Dataset $\mathcal{D}^0$ with dataset size $n$, $\ell$, Lipschitz constant of Hessian  $\beta$, gradient bound $C$, $\epsilon$, $\delta$, $\lambda$, $\boldsymbol{w}^0$, $\boldsymbol{H}^0$, $\boldsymbol{v}^1$, data value assessment method $\mathcal{S}$, the total number of deletion rounds $T$, and $\mathcal{M}^1, \mathcal{M}^2, \ldots, \mathcal{M}^T$.
\IndentedAlgorithm{
\Algphase{\texttt{DVWU} Mechanism} 
      \For{$t=1,2, \ldots,T$}       
        \State $\begin{aligned}
        \nabla ^t \gets \frac{1}{m} \sum_{i \, : \, v_i^t \neq 0} v_i^t (\nabla \ell\left(\boldsymbol{w}^{t-1}; z_i^t\right)+\lambda\boldsymbol{w}^{t-1}).\end{aligned}$\label{deltac}
        \State $\begin{aligned}
       \boldsymbol{H}^t \gets \frac{1}{n-tm} \Bigl( (n-tm+m) \boldsymbol{H}^{t-1} - \sum_{i=1}^m \left( \nabla^2 \ell\left(\boldsymbol{w}^{t-1},z_i^t\right)+\lambda \right)
      \Bigr).
       \end{aligned}$

        \State $\boldsymbol{w}^t \gets \boldsymbol{w}^{t-1} + \frac{m}{n-tm}[\boldsymbol{H}^t]^{-1} \nabla ^t$. 
        \State Update dataset: $\mathcal{D}^t \gets \mathcal{D}^{t-1} \setminus \mathcal{M}^t$.  
        \State Recalculate data values: $\boldsymbol{q}^{t+1} \gets \mathcal{S}(\mathcal{D}^{t})$.
        \State Recalculate weights: $\boldsymbol{v}^{t+1} \gets f(\boldsymbol{q}^{t+1})$.  
       
\Algphase{Certified Unlearning Guarantee} 
      \State $\boldsymbol{g}^{t} \gets \nabla L(\boldsymbol{w}^t ; \mathcal{D}^t).$\textcolor{blue}{\Comment{Not Necessary for Every Deleting}}      
     \State Draw a noise vector $\boldsymbol{b}^t \sim \mathcal{N}\left(0, (c \epsilon_1^{\prime} / \epsilon)^2 I_d\right)$, where $c=\sqrt{2\ln(1.25/\delta)}$ and  $\epsilon_1^{\prime}=\frac{4 \beta C^2 m^2 t}{\lambda^3(n-tm)^2}+\frac{4 C m t}{\lambda(n-tm)}$.\label{noisecal}    
     \If{$\left\|\boldsymbol{g}^{t}\right\|_2\leq\lambda \epsilon_1^{\prime}$}
        \State $\boldsymbol{w}^t$ is kept for the next iteration.
        \State Publish ${\boldsymbol{w}^{\prime}}^t = \boldsymbol{w}^t + \boldsymbol{b}^t$. \textcolor{blue}{\Comment{For Public Output}}
        \Else
        \State Retrain model: ${\boldsymbol{w}^*}^t \gets  \arg \min_{\boldsymbol{w}} L(\boldsymbol{w}; \mathcal{D}^t)$.
        \State  For the next iteration: ${\boldsymbol{w}}^t \gets {\boldsymbol{w}^*}^t$.
        \State Reinitialize: $\boldsymbol{q}^{t+1}, \boldsymbol{v}^{t+1}$.
        \EndIf 
     \EndFor
     }
\end{algorithmic}
\end{algorithm}

\subsection{\texttt{DVWU} with Objective Perturbation} \label{Objp}

    \texttt{DVWU} with objective perturbation is presented in Algorithm \ref{alg:DVMU2}.    
    Unlike in Algorithm \ref{alg:DVMU1}, the initial ML model is trained by optimizing the following perturbed objective function, $L^{\boldsymbol{b}}(\boldsymbol{w} ; \mathcal{D})$:

    \begin{equation}
    L^{\boldsymbol{b}}(\boldsymbol{w} ; \mathcal{D})=L(\boldsymbol{w} ; \mathcal{D})+\boldsymbol{b}^{\top} \boldsymbol{w},
    \label{eq:lossnoise}
    \end{equation}    
    where the perturbation vector $\boldsymbol{b}$ is drawn from a normal distribution, $\boldsymbol{b}\sim \mathcal{N}\left(0, (c \epsilon_2^{\prime} / \epsilon)^2 I_d\right)$. The terms are defined as $c=\sqrt{2\ln(1.25/\delta)}$ and $\epsilon_2^{\prime}=\frac{4 \beta C^2 m^2 T}{\lambda^2(n - Tm)^2}+\frac{4 C m T}{n - Tm}$, with $T$ being the total number of deletion rounds.
        
    The optimal model parameter $\boldsymbol{w}^*= \arg \min_{\boldsymbol{w}} L^{\boldsymbol{b}}(\boldsymbol{w} ; \mathcal{D})$ is saved as $\boldsymbol{w}^0$, and the Hessian of $L^{\boldsymbol{b}}$ at $\boldsymbol{w}^*$ is saved as $\boldsymbol{H}^0$, with $\boldsymbol{b}$ also saved. 
    Then, the data value-weighted unlearning proceeds.

    Theorem \ref{theorem_cob} then provides certified unlearning guarantees for Algorithm \ref{alg:DVMU2} (see proof in Appendix A.5).

\begin{theorem}\label{theorem_cob}
    \textit{Let $\mathcal{A}$ be the learning algorithm that returns the unique optimum of  $L^{\boldsymbol{b}}(\boldsymbol{w}; \mathcal{D}^t)$, then Algorithm \ref{alg:DVMU2}
     provides $(\epsilon, \delta)$-certified unlearning of $\mathcal{A}$.}
\end{theorem}

\begin{algorithm}[!ht]
\caption{\texttt{DVWU} with Objective Perturbation in Continuous Deletion}\label{alg:DVMU2}
\begin{algorithmic}[1]
\Require Dataset $\mathcal{D}^0$ with dataset size $n$, $\ell$, $\epsilon_2^{\prime}$, $\boldsymbol{w}^0$, $\boldsymbol{H}^0$, $\boldsymbol{b}$, $\boldsymbol{v}^1$, data value assessment method $\mathcal{S}$,  the total number of deletion rounds $T$, and $\mathcal{M}^1,\mathcal{M}^2,\ldots\mathcal{M}^T$.
\IndentedAlgorithm{
\Algphase{\texttt{DVWU} Mechanism}
        \For{$t=1,2, \ldots,T$}     
        \State $\begin{aligned}
        \nabla ^t \gets  \frac{1}{m} \sum_{i \, : \, v_i^t \neq 0} v_i^t (\nabla \ell\left(\boldsymbol{w}^{t-1}; z_i^t\right) +\lambda\boldsymbol{w}^{t-1}+\boldsymbol{b}).\end{aligned}$      
        \State $\begin{aligned}
       \boldsymbol{H}^t \gets \frac{1}{n-tm} \Bigl((n-tm+m) \boldsymbol{H}^{t-1} - \sum_{i=1}^m \left( \nabla^2 \ell\left(\boldsymbol{w}^{t-1},z_i^t\right)+\lambda \right)
      \Bigr).
       \end{aligned}$

        \State $\boldsymbol{w}^t \gets \boldsymbol{w}^{t-1} + \frac{m}{n-tm}[\boldsymbol{H}^t]^{-1} \nabla ^t$.          
        \State Update dataset: $\mathcal{D}^t \gets \mathcal{D}^{t-1} \setminus \mathcal{M}^t$.  
        \State Recalculate data values: $\boldsymbol{q}^{t+1} \gets \mathcal{S}(\mathcal{D}^{t})$.
        \State Recalculate weights: $\boldsymbol{v}^{t+1} \gets f(\boldsymbol{q}^{t+1})$.
      
\Algphase{Certified Unlearning Guarantee} 
  
    \State $\boldsymbol{g}^{t} \gets \nabla L^{\boldsymbol{b}}(\boldsymbol{w}^t ; \mathcal{D}^t).$\textcolor{blue}{\Comment{Not Necessary for Every Deleting}}
        \If{$\left\|\boldsymbol{g}^{t}\right\|_2\leq \epsilon_2^{\prime}$}
        \State  $\boldsymbol{w}^t$ is kept for the next iteration.
        \Else  
        \State Retrain model: ${\boldsymbol{w}^*}^t \gets  \arg \min_{\boldsymbol{w}} L^{\boldsymbol{b}}(\boldsymbol{w}; \mathcal{D}^t)$.
        \State  For the next iteration: ${\boldsymbol{w}}^t \gets {\boldsymbol{w}^*}^t$.
        \State Reinitialize: $\boldsymbol{q}^{t+1}, \boldsymbol{v}^{t+1}$.
        \EndIf 
     \EndFor
     }
         
\end{algorithmic}
\end{algorithm}

Note that Algorithm~\ref{alg:DVMU1} outputs both the original updated model parameters and the perturbed ones.
  In contrast, Algorithm~\ref{alg:DVMU2} outputs only the perturbed model parameters, as noise is injected directly into the objective function at the beginning of training.
  Therefore, since subsequent unlearning parameter updates can only be performed based on noisy parameters, the model's predictive performance may degrade.  
  However, this design offers further advantages in privacy protection.
  By introducing perturbations into the objective function, Algorithm \ref{alg:DVMU2} not only prevents leakage of gradient information during updates  \cite{ChuanGuo2019CertifiedDR}, but also enhances the model robustness against data fluctuations   \cite{pinot2019unified, lecuyer2019certified}. 
  
\begin{remark}
The main computational tasks in \texttt{DVWU} are divided into three parts: computing the weighted gradients, calculating and inverting the Hessian matrix, and evaluating data values followed by updating the data value weights. 
The complexity of computing the weighted gradients is $O(md)$. This cost can be further reduced if many of the data value weights are zero.
The Hessian-related computation, which mirrors that in the one-step Newton update, has a complexity of $O(d^3)$. 
Existing acceleration techniques--such as approximating the Hessian with the Fisher Information Matrix (FIM) \citep{AlexandraPeste2021SSSEEE} or using stochastic estimation methods \citep{NamanAgarwal2016SecondOS}--can significantly reduce this cost.
Since the focus of this paper is on the performance improvements introduced by data value weights, these acceleration techniques are not further discussed.
The complexity of computing data values depends on the specific estimation method used. 
 For instance, with HSV-KNN~\citep{zhang2023dynamic}, the complexity is $O(m \log (m) d (k+1))$, where $k$ is the number of neighbors in KNN. 
Finally, updating the data value weights requires $O(n-m)$. 
If static weights are used, the computation cost in this part can be omitted.
\end{remark} 

\section{Experiment Study}\label{sec_experiment}

This section evaluates the proposed \texttt{DVWU} framework for machine unlearning.
We begin by assessing its performance using the one-step Newton update, comparing it with baseline methods on both synthetic and real-world data sets under linear models.
We then demonstrate the framework's extensibility by applying \texttt{DVWU} to the gradient ascent method~\citep{thudi2022unrolling}, illustrating its adaptability to gradient-based deep unlearning approaches.

\subsection{Experimental Setup}

\subsubsection{Classification Models}
    \textbf{Linear Unlearning: } We apply the \texttt{DVWU} algorithms to two classification models:
    Logistic Regression (LR) and Huberized SVM (with $\gamma =2$ \citep{wang2008hybrid}).
    Furthermore, the L-BFGS-B \cite{RichardHByrd1995ALM} method is applied to perform the convex minimization procedure and obtain both the initial and retrained model parameters.

    \textbf{Deep Unlearning:} We conduct experiments on the \texttt{MNIST} image benchmark dataset \cite{lecun1998gradient} and the \texttt{SST-2} text sentiment analysis dataset \cite{socher2013recursive}.
    For the \texttt{MNIST} dataset, we employ a lightweight Convolutional Neural Network (CNN). For the \texttt{SST-2} dataset, we first encode the text into feature vectors using a pre-trained DistilBERT model \cite{Sanh2019DistilBERTAD}, which are then fed into a two-layer feed-forward network for classification. Both models are optimized using the Adam optimizer \cite{kingma2014adam}. Detailed model architectures and hyperparameter configurations are provided in Appendix B.8.

 \subsubsection{Baselines}

    \textbf{Linear Unlearning: } We consider LR and SVM and test the unlearning performance for each of them. 
    To evaluate our \texttt{DVWU} approach, we establish several baselines for each classifier, including full model retraining, the unweighted one-step Newton update \cite{ChuanGuo2019CertifiedDR}, the unweighted influence method \cite{PangWeiKoh2017UnderstandingBP,RyanGiordano2018ASA}, and the unweighted gradient ascent \cite{thudi2022unrolling}.    
    Four implementations of \texttt{DVWU} are developed for each classification model, which utilize the KNN-SV \cite{Jia_DaoETSDV2019} and the leave-one-out \cite{RDennisCook1977DetectionOI} separately for static data value assessment, and the HSV-KNN~\cite{zhang2023dynamic} based on the two static methods for dynamic data value assessment. Specifically, when calculating $\boldsymbol{v}^t$ from $ f(\boldsymbol{q}^t)$, we set $\alpha =0.5$.
     For clarity, the notation prefixes for all linear unlearning methods are described in Table \ref{tab:Alnotations}. 
    
    \textbf{Deep Unlearning:} For deep learning models, calculating the Hessian matrix for the one-step Newton update is computationally prohibitive. 
    Thus, we use retraining and gradient ascent \cite{thudi2022unrolling} as our primary baselines. 
    We compare them against our proposed data value-weighted gradient ascent (see details in Appendix B.9), which we evaluate using two weighting strategies: static weights and dynamic weights. For data valuation in this context, we exclusively use the KNN-SV \cite{Jia_DaoETSDV2019} method.
    Detailed  descriptions of classifiers and methods are provided in Appendix B.10.

\begin{table*}[!ht]
\caption{Notations for  Methods in Linear Unlearning}\label{tab:Alnotations}
\centering
\scriptsize
\begin{tabular}{l c r}
\hline \textbf{Notations} & \textbf{ Unlearning Method} & \textbf{Data Value Weight}\\
\hline \multicolumn{3}{l}{Baseline Methods} \\
\hline \texttt{Retrain} & Retraining & No Weight \\
 \texttt{Newton} \cite{ChuanGuo2019CertifiedDR} & One-Step Newton Update & No Weight \\
\texttt{Influence} \cite{PangWeiKoh2017UnderstandingBP,RyanGiordano2018ASA} & Influence Method & No Weight \\ \texttt{GradientA} \cite{thudi2022unrolling} & Gradient Ascent & No Weight \\
\hline \multicolumn{3}{l}{DVWU} \\
\hline \texttt{DVWUk} & Static Weight One-Step Newton Update & KNN-SV \cite{Jia_DaoETSDV2019} \\
 \texttt{DVWUl} & Static Weight One-Step Newton Update  & Leave-One-Out \cite{RDennisCook1977DetectionOI} \\
 \texttt{DVWUdk} & Dynamic Weight One-Step Newton Update & KNN-SV \cite{Jia_DaoETSDV2019}, HSV-KNN 
 \cite{zhang2023dynamic} \\
 \texttt{DVWUdl} & Dynamic Weight One-Step Newton Update & Leave-One-Out \cite{RDennisCook1977DetectionOI}, HSV-KNN \cite{zhang2023dynamic} \\
\hline
\end{tabular}
\end{table*}

\subsubsection{Performance Evaluation}

    We assess model performance across four dimensions: \textit{unlearning efficiency}, \textit{unlearning capability}, \textit{unlearning certifiability}, and \textit{unlearning robustness}. 
    Each dimension is detailed below:

    \begin{itemize}
    \item \textit{Unlearning efficiency} is assessed by an experiment in which random $1,000$ data points are deleted simultaneously. 
    We compare the running time of all unlearning methods to that of retraining.
    
    \item \textit{Unlearning capability} is evaluated through a continuous deletion experiment where  $1,000$ random data points are deleted in each round, over a total of $15$ rounds. 
    We evaluate several metrics for unlearning methods, including accuracy, precision, recall, and misclassification cost. 
     
    \item \textit{Unlearning certifiability} is verified by computing gradient residuals of unlearning methods at each round.
    
    \item \textit{Unlearning robustness} is evaluated through experiments on datasets with varying noise levels and on imbalanced datasets. 
    We calculate the same performance metrics as those in \textit{unlearning capability}.
    \end{itemize}

All experimental results presented are averages over $100$ independent runs.

\subsection{Results of Linear Unlearning}
\subsubsection{Synthetic Datasets} 

\paragraph{Data Generation}\label{symethod}
We first generate the dataset by randomly placing two Gaussian clusters for class -1 and two Gaussian clusters for class 1 at the vertices of a two-dimensional hypercube with side lengths of 2. 
   For each cluster, the features are independently drawn from a standard normal distribution $\mathcal{N}(0, 1)$.
   To increase the complexity of the data, we create two more redundant features by linearly combining the original features.
   We then introduce additional noise by randomly flipping 10\% of the class labels. 
   This method adapts from  \cite{YiYang2021PrivacyPreservingCL,guyon2003design}, and the generated dataset is used in our motivating example. 

   We also generate six additional synthetic data sets using the same method, with detailed specifications provided in Table~\ref{tab:syDatainstruction}.
   Among these, \texttt{sy1} is used to assess unlearning capability and unlearning certifiability.
   \texttt{sy1}, \texttt{sy2}, and \texttt{sy3}, which differ only in noise ratios, are employed to evaluate unlearning robustness. 
   \texttt{sy1}, \texttt{sy4}, and \texttt{sy5} are used to assess unlearning efficiency across different data scales.
   Finally, \texttt{sy1} and \texttt{sy6}, which differ only in their positive sample ratio, are used to investigate the effect of imbalanced samples on unlearning robustness.
   All datasets are split into training and testing sets at a 7:3 ratio.

\begin{table}[H]
\caption{\label{tab:syDatainstruction}Summary of Synthetic Data}
\centering
\begin{tabular}{l c c c r}
\hline  \textbf{Data} & \textbf{$n$} & \textbf{$d$} & \textbf{Positive Ratio} & \textbf{Noise Ratio} \\
\hline \texttt{sy1} & 30000 & 20 & 0.5 & 0.05 \\
 \hline \texttt{sy2} & 30000 & 20 & 0.5 & 0.15 \\
\hline \texttt{sy3} & 30000 & 20 & 0.5 & 0.25 \\
\hline \texttt{sy4} & 30000 & 40 & 0.5 & 0.05 \\
 \hline \texttt{sy5} & 60000 & 40 & 0.5 & 0.05 \\
\hline \texttt{sy6} & 30000 & 20 & 0.25 & 0.05 \\
\hline
\end{tabular}
\end{table}

\paragraph{Unlearning Capability}\label{dvcompare}
We compare the performance of all  methods in a continuous deletion scenario on data \texttt{sy1}. 
Figure \ref{fig:sy1lr1} shows the LR results with our output perturbation method. We can find that as the number of deleted data points increases, the performances of the baselines—\texttt{Retrain}, \texttt{Newton}, \texttt{Influence}, and \texttt{GradientA}—gradually decline.
This result confirms that directly applying unlearning algorithms for model updates can lead to a degradation in model performance, especially as the amount of deleted data accumulates.
In contrast, \texttt{DVWUk} and \texttt{DVWUdk} demonstrate enhanced performance, showing improvements in accuracy and precision. Additionally, \texttt{DVWUl} and \texttt{DVWUdl} show inferior performance compared to the other approaches. 
This difference is rooted in their distinct data valuation methods.
\texttt{DVWUl} and \texttt{DVWUdl} use the leave-one-out method to estimate data value, but its accuracy degrades as more data is deleted, limiting its effectiveness for unlearning. In contrast, \texttt{DVWUk} and \texttt{DVWUdk} leverages KNN-SV, which computes values based on sample distances to the test set and remains robust to data deletion, leading to superior performance.

Due to space limitations,  results for SVM and results of our objective perturbation method are provided in Appendix B.1, which show a similar trend.

\begin{figure}[H]
\centering
\includegraphics[width=\linewidth]{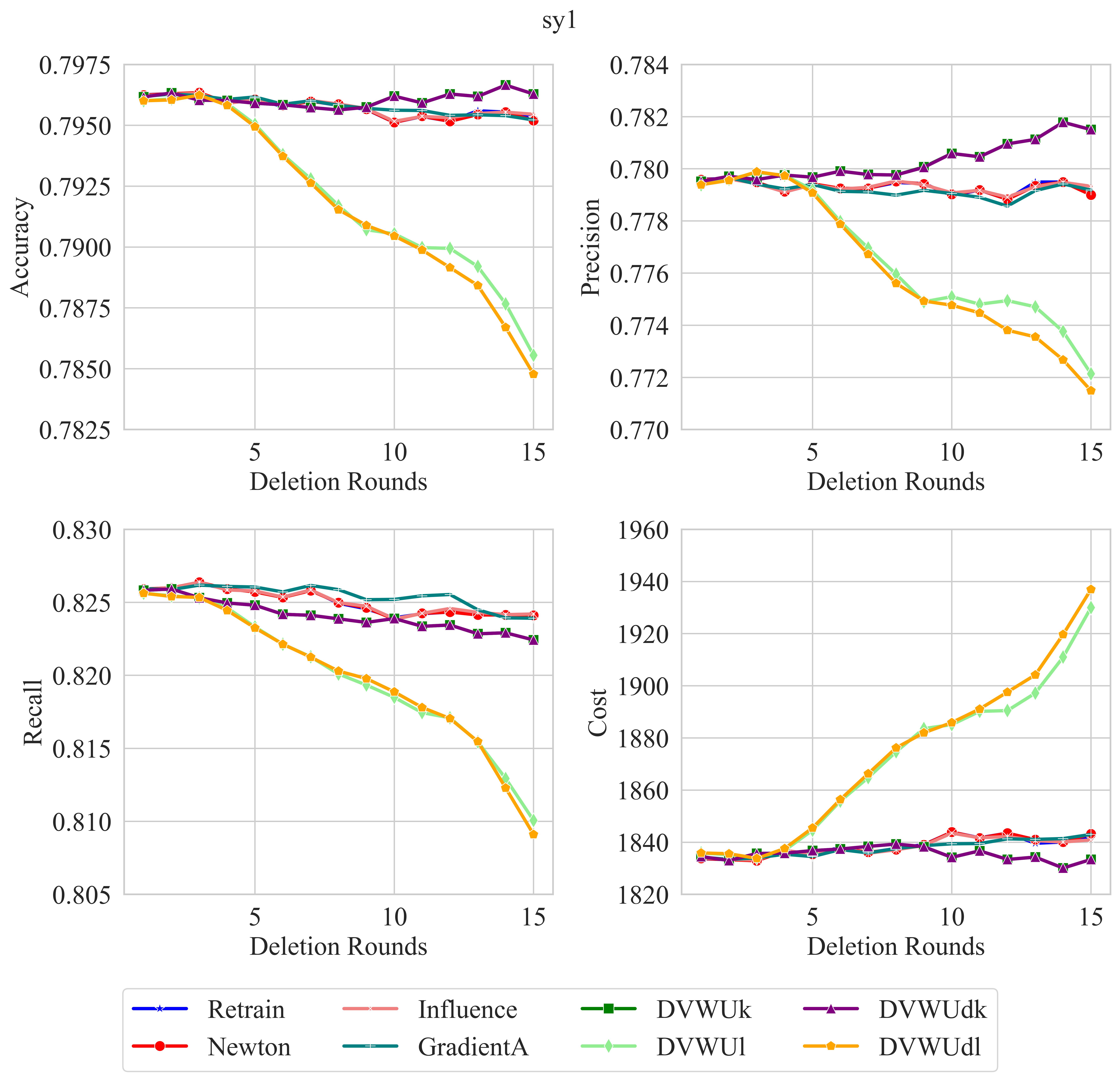}%
\caption{\label{fig:sy1lr1}Model performance of continuous deletions with output perturbation using  LR on data \texttt{sy1}.}
\end{figure}

\paragraph{Unlearning Certifiability}\label{uncertifia} 
    To verify unlearning certifiability, on \texttt{sy1}, we set $\delta = 1 \times 10^{-4}$ and $\epsilon = 1$. With $m = 1000$ and $\lambda = 0.001$, the certifiability thresholds can be calculated as in Algorithms \ref{alg:DVMU1} and \ref{alg:DVMU2}. 
   Here, we consider two scenarios: \textbf{Threshold 0} for the specific case where $v_i^t = 0$ for all $i$, and \textbf{Threshold 1} for the general case without assumption on $v_i^t$.
   When using output perturbation, the threshold is $\lambda \epsilon_1'$. 
   For Threshold 1, $\epsilon_1'$ can be computed as in Line~\ref{noisecal} of Algorithm~\ref{alg:DVMU1}.  
   For Threshold 0, the calculation can be simplified, and $\epsilon_1'$ reduces to $\frac{2 C m t}{\lambda(n - t m)}$ (see Appendix A.2).
   The threshold for objective perturbation can be computed in a similar way.
    
    Figure \ref{fig:sy1gradient} illustrates the gradient residual norms under continuous deletion for both LR and SVM models with our output perturbation method.
    The results show that the gradient residual norms of \texttt{DVWUk}, \texttt{DVWUl}, \texttt{DVWUdk}, and \texttt{DVWUdl} gradually accumulate but consistently remain below both thresholds within tolerance.
    This confirms the certifiability of \texttt{DVWU}.     
    This further indicates that, for a given perturbation threshold, our method enables the deletion of more data than the target size $m$.
    From another perspective, this also suggests the potential to reduce the injected noise, thereby mitigating performance degradation caused by perturbations.
    Note that in our algorithms, the gradient residual norm is primarily computed for certifiability checks. Since a sufficient margin below the threshold is observed, the strong certifiability guarantee holds, making it unnecessary to compute the norm at every deletion round.    
   
    Due to space limitations, results for our objective perturbation method are provided in Appendix B.2, showing a similar trend.
   
    \begin{figure}[t]
        \centering
        \includegraphics[width=\linewidth]{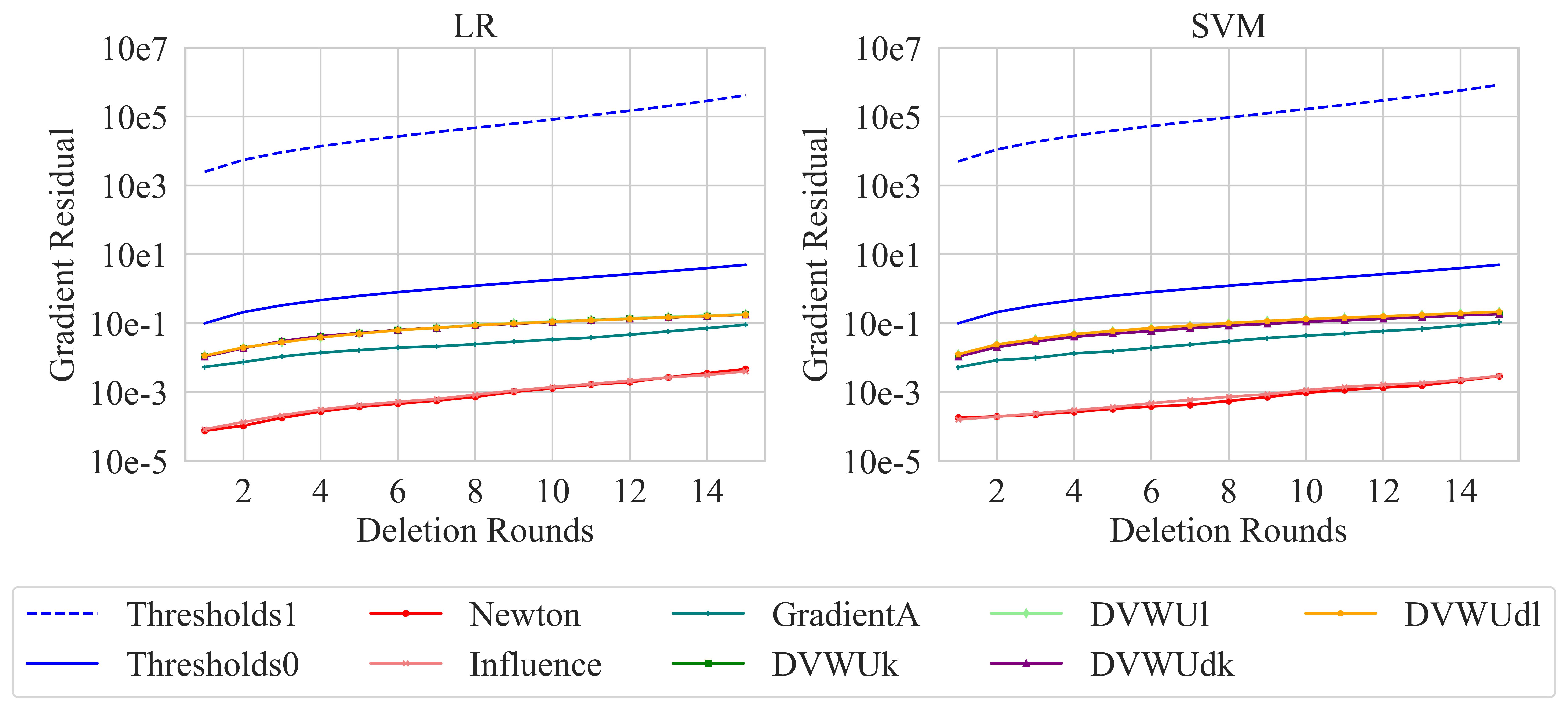}
        \caption{Unlearning certifiability of  continuous deletions with output perturbation on data \texttt{sy1}.}
        \label{fig:sy1gradient}
    \end{figure}

\paragraph{Unlearning Efficiency}
The running time of all methods on data \texttt{sy1}, \texttt{sy4}, and \texttt{sy5} is shown in Table \ref{tab:syEfficiency}. 
In all cases, \texttt{Retrain} is the most time-consuming method, while unlearning approaches, including \texttt{DVWUl}, \texttt{DVWUk}, \texttt{Newton}, \texttt{Influence} and \texttt{GradientA}, are significantly faster. 
\texttt{Influence} achieves the relatively lower runtime because it pre-computes the inverse of the Hessian matrix; during unlearning, the main computations are limited to evaluating the gradient and multiplying it by the pre-computed inverse Hessian, which is more efficient than the operations required by \texttt{Retrain}, \texttt{DVWUl}, \texttt{DVWUk}, and \texttt{Newton}.
Since \texttt{GradientA} primarily computes the gradient, it has the lowest runtime.
As the dataset size increases, the time required for \texttt{Retrain} grows substantially, whereas the times for \texttt{DVWUl}, \texttt{DVWUk}, \texttt{Newton}, \texttt{Influence}, and \texttt{GradientA} remain relatively stable, underscoring the efficiency advantage of  unlearning.
Compared to \texttt{Newton}, \texttt{DVWUl} and \texttt{DVWUk} require slightly more time, primarily due to the additional multiplication operations involved in gradient weighting, which offset the time saved by skipping gradient calculations for points with zero value.
However, the running time for \texttt{DVWUl} and \texttt{DVWUk} remains within the same order of magnitude as \texttt{Newton}, demonstrating their comparable computational efficiency.

Due to space limitations, results for our objective perturbation method can be found in Appendix B.3, which shows a similar trend.

\begin{table*}[t]
\caption{\label{tab:syEfficiency}Running Time of All Methods with Output Perturbation on Synthetic Data }
\centering
\scriptsize
\begin{tabular}{l c c c c c c r}
\hline \textbf{Data} & \textbf{Classification} & \texttt{Retrain} & \texttt{Newton} & \texttt{Influence} & \texttt{GradientA} & \texttt{DVWUk} & \texttt{DVWUl} \\
\hline \texttt{sy1} & LR & 0.0176 & 0.0045 & 0.0002 & 0.0002 & 0.0053 & 0.0052 \\
\hline \texttt{sy1} & SVM & 0.3559 & 0.0046 & 0.0005 & 0.0005 & 0.0052 & 0.0052 \\
 \hline \texttt{sy4} & LR & 0.0221 & 0.0048 & 0.0002 & 0.0001 & 0.0055 & 0.0056 \\
 \hline \texttt{sy4} & SVM & 0.3734 & 0.0049 & 0.0006 & 0.0005 & 0.0056 & 0.0056 \\
\hline \texttt{sy5} & LR & 0.0426 & 0.0048 & 0.0002 & 0.0001 & 0.0052 & 0.0056 \\
\hline\texttt{sy5} & SVM & 0.7901 & 0.0048 & 0.0005 & 0.0005 & 0.0053 & 0.0056 \\
\hline
\end{tabular}
\end{table*}

\paragraph{Unlearning Robustness} 
    Figure~\ref{fig:sy23noi} presents the model accuracy across datasets with varying noise ratios when applying our output perturbation method. 
    The results in Figures~\ref{fig:sy1lr1} and~\ref{fig:sy23noi} show that as the noise ratio increases, the overall performance of all models deteriorates. However, \texttt{DVWUk} and \texttt{DVWUdk} consistently outperform   baselines.
    This demonstrates that \texttt{DVWU} maintains robust performance under different noise ratios. 
    Additional experiments for objective perturbation show consistent trends, as detailed in Appendix B.4.1.

   Furthermore, to assess model robustness on imbalanced data, we evaluate all methods on the \texttt{sy6} data set.
   Due to space limitations, the complete results are provided in Appendix B.4.2.
   These results further confirm that our method achieves superior performance on imbalanced data.

    \begin{figure}[t]
        \centering
        \includegraphics[width=\linewidth]{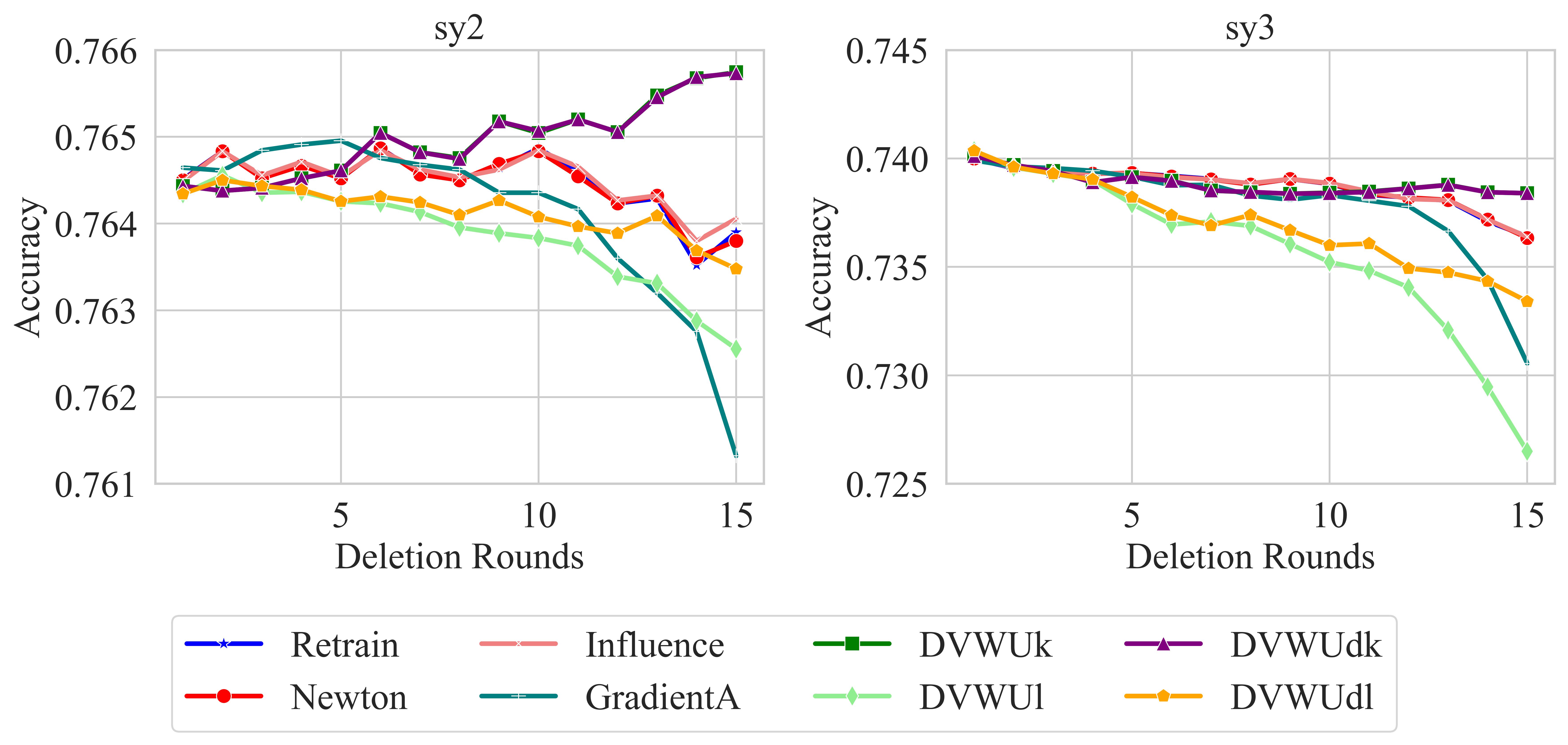}
        \caption{Model performance of  continuous deletions with output perturbation using LR on \texttt{sy2} and \texttt{sy3}.}
        \label{fig:sy23noi}
    \end{figure}

\subsubsection{Real-World Datasets}

\paragraph{Dataset Details}
    We conduct empirical validation on two real-world credit risk assessment datasets: \texttt{GMC} and \texttt{DCCC}. 
    The \texttt{GMC} (Give Me Some Credit) dataset was introduced in a Kaggle competition \cite{StefanLessmann2015BenchmarkingSC}, while the \texttt{DCCC} (Default of Credit Card Clients) dataset is publicly available in the UCI ML Repository \cite{Yeh2007}. 
    We standardize both datasets and split them into training and testing sets with a $7:3$ ratio. 
    More details about datasets are  presented in Appendix B.5.

\paragraph{Results} 
   Figure \ref{fig:GMClrc} presents results with our output perturbation method on \texttt{GMC}.    
    The results  show that when using LR and SVM, \texttt{DVWUdk} and \texttt{DVWUk} achieve higher accuracy than the baseline methods. In contrast, \texttt{DVWUdl} and \texttt{DVWUl} perform comparably to the baselines when using SVM.
    These findings indicate that our method exhibits similar performance on real datasets, demonstrating its generalizability. 

    Due to space limits, full results for our objective perturbation method on \texttt{GMC} and for both output and objective perturbations on \texttt{DCCC} are provided in Appendix B.6. 
    Unlearning efficiency results on real-world data are available in Appendix B.7.
    
    \begin{figure}[t]
        \centering
    \includegraphics[width=\linewidth]{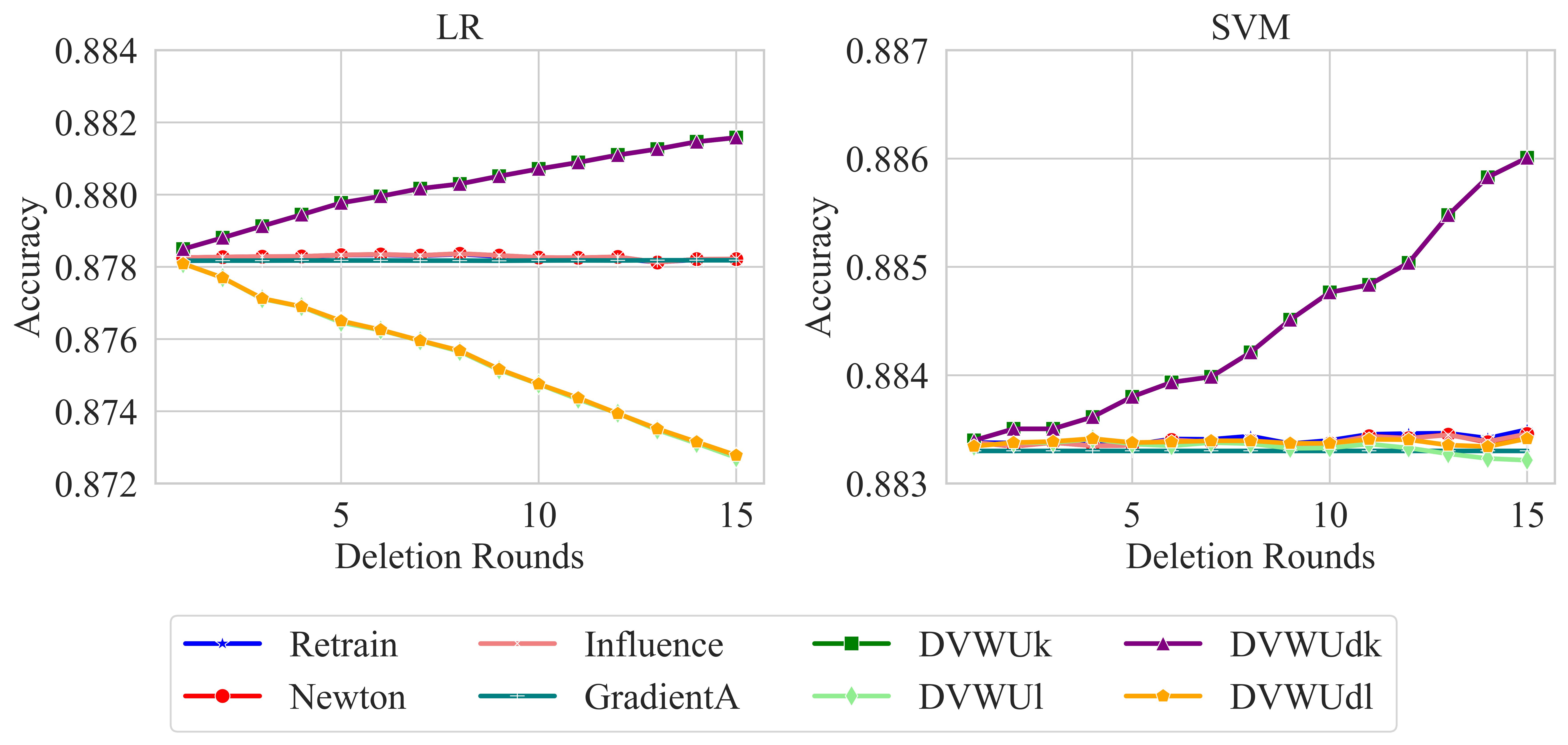}
        \caption{Model performance of  continuous deletions with output perturbation using LR and SVM on data \texttt{GMC}.}
        \label{fig:GMClrc}
    \end{figure}

\subsection{Results of Deep Unlearning}\label{nnexp}

Figure \ref{fig:nn} shows results on \texttt{MNIST} and \texttt{SST2} datasets for deep unlearning.
It clearly shows a significant and continuous decline in performance for \texttt{GradientA} as the round of deletions increases.
In contrast, our two proposed  methods not only maintain  stable model performance but also achieve slightly higher accuracy than \texttt{Retrain} in certain cases.
This observation aligns with our findings on linear models, demonstrating that the \texttt{DVWU} framework can be effectively generalized to gradient-based deep unlearning methods, thus confirming its strong generalizability and scalability.

Furthermore, we evaluate the computational efficiency of different deep unlearning methods, with details provided in Appendix B.12. The results show that our methods require substantially less time than retraining while remaining comparable to \texttt{GradientA}, highlighting their practical efficiency.

    \begin{figure}[t]
        \centering
    \includegraphics[width=\linewidth]{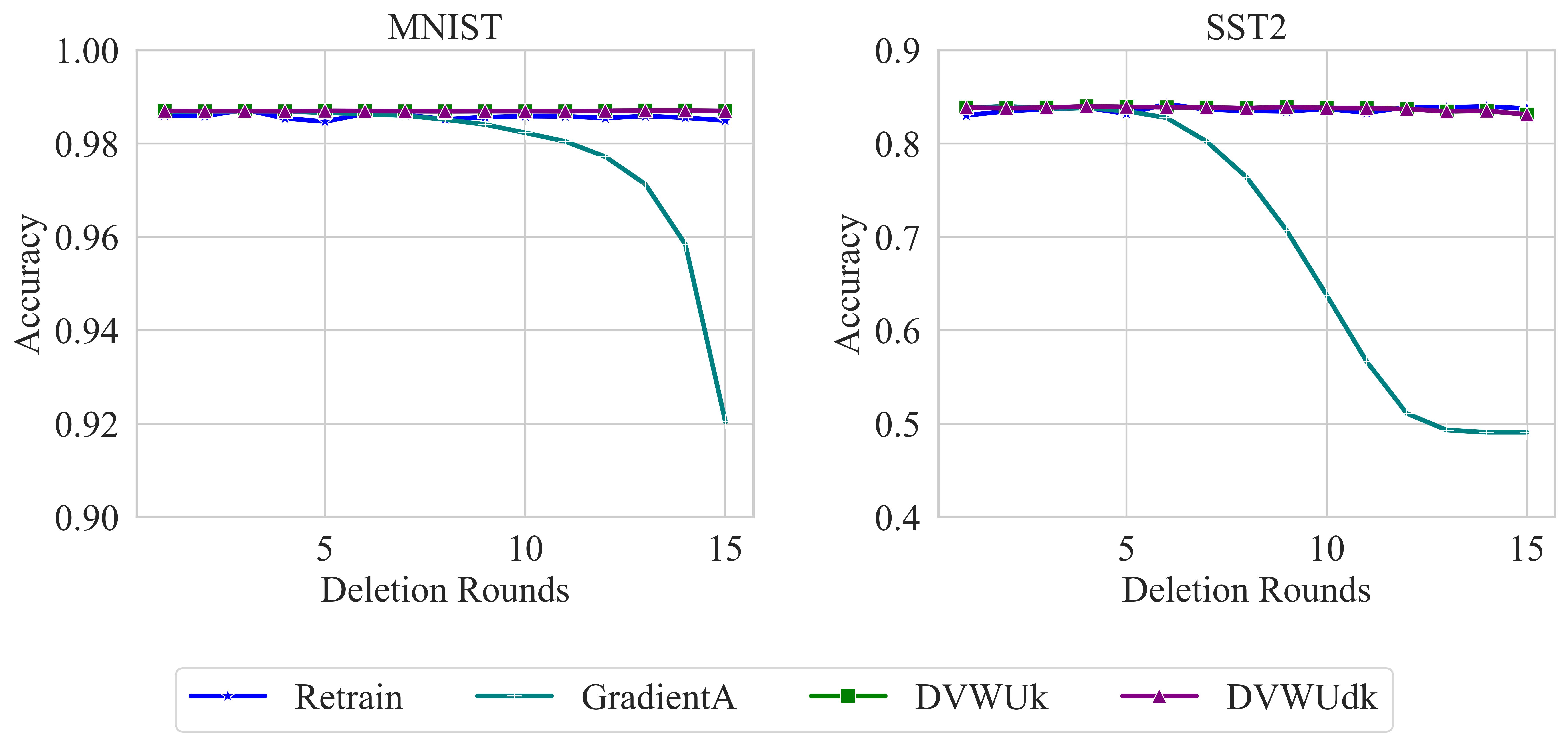}
        \caption{Model performance of continuous deletions on data \texttt{MNIST} and \texttt{SST2}.}
        \label{fig:nn}
    \end{figure}

\section{Conclusion}\label{sec_conclusion}
    
    Since individual data points contribute unequally to model performance, the impact of their deletion naturally varies. However, previous unlearning methods overlook this heterogeneity, which results in performance degradation of the updated model.  
    To address this, we propose \texttt{DVWU}, a novel framework that incorporates data value heterogeneity into the unlearning process, thereby better preserving model accuracy after data deletion.
   We first develop a data value weighting strategy and then integrate it into the unlearning procedure to account for the importance of the data being removed, resulting in the \texttt{DVWU} framework.
    For implementation, we use the one-step Newton update as an example and propose two \texttt{DVWU} algorithms--based on output perturbation and objective perturbation--to guarantee certified unlearning.
    Experimental results show that \texttt{DVWU} outperforms conventional approaches in maintaining model performance and robustness, while remaining computationally efficient. 
    Furthermore, we show that the proposed method is broadly applicable and can be  extended to gradient-based deep unlearning methods. 
    These results highlight the advantages of \texttt{DVWU} as a general framework for value-aware model updates in machine unlearning.
    
   Despite its advantages, the proposed method has several limitations.
    First, our method has not yet been evaluated in the context of machine unlearning for large language models (LLMs).    
    LLMs typically involve billions of parameters, making operations such as gradient computation, Hessian approximation, and data value estimation highly memory-intensive and computationally expensive. Therefore, a key direction for future work is to adapt the DVWU framework to unlearning methods for LLMs~\cite{maini2024tofu,cha2024learning} by designing efficient and memory-conscious variants. 
    Such extensions could enable efficient unlearning in LLMs while preserving model performance.    
        
    Second, the computational efficiency and accuracy of the proposed \texttt{DVWU} framework are significantly influenced by the method used for data value computation. Selecting an appropriate data value assessment method is crucial for maintaining both model performance and computational efficiency. In this paper, we consider three widely adopted data value assessment methods. Future research could explore more advanced techniques--such as those based on ensemble learning or probabilistic modeling--to further enhance the effectiveness of \texttt{DVWU}.


\begin{thebibliography}{1}
\scriptsize
\bibitem{MagorzataMagdziarczyk2019RightTB}
M.~Magdziarczyk, ``Right to be forgotten in light of regulation (eu) 2016/679 of the european parliament and of the council of 27 april 2016 on the protection of natural persons with regard to the processing of personal data and on the free movement of such data and repealing directive 95/46/ec,'' 2019.

\bibitem{California}
S.~L. Pardau, ``The california consumer privacy act: Towards a european-style privacy regime in the united states,'' \textit{J. Tech. L. \& Pol'y}, vol.~23, p.~68, 2018.


\bibitem{MinChen2021WhenMU}
M.~Chen, Z.~Zhang, T.~Wang, M.~Backes, M.~Humbert, and Y.~Zhang, ``When machine unlearning jeopardizes privacy,'' in \textit{Proceedings of the 2021 ACM SIGSAC Conference on Computer and Communications Security}, 2021, pp. 896--911.

\bibitem{YinzhiCao2015TowardsMS}
Y.~Cao and J.~Yang, ``Towards making systems forget with machine unlearning,'' in \textit{2015 IEEE Symposium on Security and Privacy}, 2015, pp. 463--480.

\bibitem{LucasBourtoule2019MachineU}
L.~Bourtoule, V.~Chandrasekaran, C.~A. Choquette-Choo, H.~Jia, A.~Travers, B.~Zhang, D.~Lie, and N.~Papernot, ``Machine unlearning,'' \textit{2021 IEEE Symposium on Security and Privacy (SP)}, 
 2019, pp. 141--159.

\bibitem{AdityaGolatkar2020MixedPrivacyFI}
A.~Golatkar, A.~Achille, A.~Ravichandran, M.~Polito, and S.~Soatto, ``Mixed-privacy forgetting in deep networks,'' \textit{2021 IEEE/CVF Conference on Computer Vision and Pattern Recognition (CVPR)}, 2021, pp. 792--801.

\bibitem{PangWeiKoh2017UnderstandingBP}
P.~W. Koh and P.~Liang, ``Understanding black-box predictions via influence functions,'' \textit{Proceedings of the 34th International Conference on Machine Learning (ICML 2017)}, 2017, pp. 1885--1894.

\bibitem{RyanGiordano2018ASA}
R.~Giordano, W.~T. Stephenson, R.~Liu, M.~I. Jordan, and T.~Broderick, ``A swiss army infinitesimal jackknife,'' in \textit{Proceedings of the Twenty-Second International Conference on Artificial Intelligence and Statistics}, 2019, pp. 1139--1147.

\bibitem{ChuanGuo2019CertifiedDR}
C.~Guo, T.~Goldstein, A.~Hannun, and L.~Van Der~Maaten, ``Certified data removal from machine learning models,'' in \textit{Proceedings of the 37th International Conference on Machine Learning}, 2020, pp. 3832--3842. 

\bibitem{AyushKTarun2021FastYE}
A.~K. Tarun, V.~S. Chundawat, M.~Mandal, and M.~S. Kankanhalli, ``Fast yet effective machine unlearning,'' \textit{IEEE transactions on neural networks and learning systems}, 2023, pp. 13046-13055.

\bibitem{Schelter21He}
S.~Schelter, S.~Grafberger, and T.~Dunning, ``Hedgecut: Maintaining randomised trees for low-latency machine unlearning,'' \textit{Proceedings of the 2021 International Conference on Management of Data}, 2021, pp. 1545-1557.

\bibitem{pawelczyk2022trade}
M.~Pawelczyk, T.~Leemann, A.~Biega, and G.~Kasneci, ``On the trade-off between actionable explanations and the right to be forgotten,'' \textit{arXiv preprint arXiv: 2208.14137}, 2022.

\bibitem{Tian2024DeRDaVa}
X.~Tian, R.~H.~L. Sim, J.~Fan, and B.~K.~H. Low, ``Derdava: Deletion-robust data valuation for machine learning,'' in \textit{Proceedings of the AAAI Conference on Artificial Intelligence}, vol.~38, no.~14, pp. 15\,373--15\,381, 2024.

\bibitem{RDennisCook1977DetectionOI}
R.~D. Cook, ``Detection of influential observation in linear regression,'' \textit{Technometrics}, vol.~19, no.~1, pp.~15--18, 1977.
            
\bibitem{Jia_DaoETSDV2019}
R.~Jia, D.~Dao, B.~Wang, F.~A. Hubis, N.~M. Gurel, B.~Li, C.~Zhang, C.~Spanos, and D.~Song, ``{Efficient task-specific data valuation for nearest neighbor algorithms},'' \textit{Proceedings of the VLDB Endowment},2019, pp. 1610–1623. 

\bibitem{zhang2023dynamic}
J.~Zhang, H.~Xia, Q.~Sun, J.~Liu, L.~Xiong, J.~Pei, and K.~Ren, ``Dynamic shapley value computation,'' in \textit{2023 IEEE 39th International Conference on Data Engineering (ICDE)}. 2023, pp. 639--652.

\bibitem{SalvatoreMercuri2022AnIT}
S.~Mercuri, R.~Khraishi, R.~Okhrati, D.~Batra, C.~Hamill, T.~Ghasempour, and A.~Nowlan, ``An introduction to machine unlearning,'' \textit{arXiv preprint arXiv: 2209.00939}, 2022.

\bibitem{JonathanBrophy2020MachineUF}
J.~Brophy and D.~Lowd, ``Machine unlearning for random forests,'' 2021, pp. 1092-1104.

\bibitem{YinjunWu2020DeltaGradRR}
Y.~Wu, E.~Dobriban, and S.~B. Davidson, ``Deltagrad: Rapid retraining of machine learning models,'' in \textit{Proceedings of the 37th International Conference on Machine Learning}, 2020, pp. 10\,355--10\,366.

\bibitem{YangLiu2023BackdoorDW}
Y.~Liu, M.~Fan, C.~Chen, X.~Liu, Z.~Ma, L.~Wang, and J.~Ma, ``Backdoor defense with machine unlearning,'' \textit{IEEE INFOCOM 2022 - IEEE Conference on Computer Communications}, 2022, pp. 280--289.

\bibitem{Sekhari2021RememberWY}
A.~Sekhari, J.~Acharya, G.~Kamath, and A.~T. Suresh, ``Remember what you want to forget: Algorithms for machine unlearning,''  in \textit{Advances in Neural Information Processing Systems}, 
vol.~34, 18\,075--18\,086, 2021.

\bibitem{EliChien2022CertifiedGU}
E.~Chien, C.~Pan, and O.~Milenkovic, ``Certified graph unlearning,''   \textit{arXiv preprint arXiv: 2206.09140}, 2022.

\bibitem{influenceoo}
L.~A. Jaeckel, ``The infinitesimal jackknife,'' Bell Labs, Murray Hill, NJ, Tech. Rep., 1972.

\bibitem{Shapley19977AV}
L.~S. Shapley, ``A value for n-person games,'' pp. 307--318, 1997. 

\bibitem{Wang_APATAD_2020}
T.~Wang, J.~Rausch, C.~Zhang, R.~Jia, and D.~Song, ``A Principled Approach to Data Valuation for Federated Learning,'' in \textit{Federated Learning: Privacy and Incentive}, Springer, 2020, pp.~153--167.

\bibitem{Lundberg_Lee_2017}
S.~M. Lundberg and S.-I. Lee, ``A unified approach to interpreting model predictions,'' in \textit{Proceedings of the 31st International Conference on Neural Information Processing Systems}, 2017, pp. 4768–4777.

\bibitem{Burgess_Chapman_2021}
M.~A. Burgess and A.~C. Chapman, ``{Approximating the shapley value using stratified empirical bernstein sampling},'' in \textit{Proceedings of the Thirtieth International Joint Conference on Artificial Intelligence}, 2021, pp. 73-81.

\bibitem{xia2023equitable}
H.~Xia, J.~Liu, J.~Lou, Z.~Qin, K.~Ren, Y.~Cao, and L.~Xiong, ``Equitable data valuation meets the right to be forgotten in model markets,'' \textit{Proceedings of the VLDB Endowment}, 2023, pp. 3349--3362.

\bibitem{thudi2022unrolling}
A.~Thudi, G.~Deza, V.~Chandrasekaran, and N.~Papernot, ``Unrolling sgd: Understanding factors influencing machine unlearning,'' in \textit{2022 IEEE 7th European Symposium on Security and Privacy (EuroS\&P)}, 2022, pp. 303--319.

\bibitem{yao2025large}
Y.~Yao, X.~Xu, and Y.~Liu, ``Large language model unlearning,'' in \textit{Advances in Neural Information Processing Systems}, vol.~37, pp. 105\,425--105\,475, 2024.

\bibitem{RichardJBeckman1974TheDO}
R.~J. Beckman and H.~J. Trussell, ``The distribution of an arbitrary studentized residual and the effects of updating in multiple regression,'' \textit{Journal of the American Statistical Association}, vol.~69, pp. 199--201, 1974.

\bibitem{Koh2019OnTA}
P.~W. Koh, K.-S. Ang, H.~Teo, and P.~S. Liang, ``On the accuracy of influence functions for measuring group effects,'' in \textit{Advances in Neural Information Processing Systems}, vol.~32, 2019. 

\bibitem{pinot2019unified}
R.~Pinot, F.~Yger, C.~Gouy-Pailler, and J.~Atif, ``A unified view on differential privacy and robustness to adversarial examples,''  \textit{arXiv preprint arXiv:1906.07982}, 2019.

\bibitem{lecuyer2019certified}
M.~Lecuyer, V.~Atlidakis, R.~Geambasu, D.~Hsu, and S.~Jana, ``Certified robustness to adversarial examples with differential privacy,'' in \textit{2019 IEEE symposium on security and privacy (SP)}, 2019, pp. 656--672.

\bibitem{AlexandraPeste2021SSSEEE}
A.~Peste, D.~Alistarh, and C.~H. Lampert, ``Ssse: Efficiently erasing samples from trained machine learning models,''  \textit{arXiv preprint arXiv: 2107.03860}, 2021.

\bibitem{NamanAgarwal2016SecondOS}
N.~Agarwal, B.~Bullins, and E.~Hazan, ``Second-order stochastic optimization for machine learning in linear time,'' \textit{Journal of Machine Learning Research}, vol.~18, no. 116, pp. 1--40, 2017. 

\bibitem{wang2008hybrid}
L.~Wang, J.~Zhu, and H.~Zou, ``Hybrid huberized support vector machines for microarray classification and gene selection,'' \textit{Bioinformatics}, vol.~24, no.~3, pp. 412--419, 2008.

\bibitem{YiYang2021PrivacyPreservingCL}
Y.~Yang, S.~Huang, W.~Huang, and X.~Chang, ``Privacy-preserving cost-sensitive learning,'' \textit{IEEE Transactions on Neural Networks and Learning Systems}, vol.~32, pp. 2105--2116, 2020.

\bibitem{RichardHByrd1995ALM}
R.~H. Byrd, P.~Lu, J.~Nocedal, and C.~Zhu, ``A limited memory algorithm for bound constrained optimization,'' \textit{SIAM Journal on Scientific Computing}, vol.~16, no.~5, pp. 1190--1208, 1995.


\bibitem{lecun1998gradient}
Y.~LeCun, L.~Bottou, Y.~Bengio, and P.~Haffner, ``Gradient-based learning applied to document recognition,'' \textit{Proceedings of the IEEE}, vol.~86, no.~11, pp. 2278--2324, 1998.

\bibitem{socher2013recursive}
R.~Socher, A.~Perelygin, J.~Wu, J.~Chuang, C.~D. Manning, A.~Y. Ng, and C.~Potts, ``Recursive deep models for semantic compositionality over a sentiment treebank,'' in \textit{Proceedings of the 2013 Conference on Empirical Methods in Natural Language Processing}, 2013, pp. 1631--1642.

\bibitem{Sanh2019DistilBERTAD}
V.~Sanh, L.~Debut, J.~Chaumond, and T.~Wolf, ``DistilBERT, a distilled version of BERT: smaller, faster, cheaper and lighter,'' \textit{arXiv preprint arXiv:  1910.01108}, 2019.

\bibitem{kingma2014adam}
D.~P. Kingma and J. Ba, ``Adam: A method for stochastic optimization,'' \textit{arXiv preprint arXiv:1412.6980}, 2014.

\bibitem{guyon2003design}
I.~Guyon, ``Design of experiments of the nips 2003 variable selection benchmark,'' in \textit{NIPS 2003 workshop on feature extraction and feature selection}, vol. ~253, pp.~40, 2003.

\bibitem{StefanLessmann2015BenchmarkingSC}
B.~Baesens, T.~V. Gestel, S.~Viaene, M.~Stepanova, J.~A.~K. Suykens, and J.~Vanthienen, ``Benchmarking state-of-the-art classification algorithms for credit scoring,'' \textit{Journal of the Operational Research Society}, vol.~54, pp. 627--635, 2003.

\bibitem{Yeh2007}
I.-C. Yeh and C.-h. Lien, ``The comparisons of data mining techniques for the predictive accuracy of probability of default of credit card clients,'' \textit{Expert Syst. Appl.}, vol.~36, no.~2, pp. 2473–2480, 2009. 

\bibitem{maini2024tofu}
P.~Maini, Z.~Feng, A.~Schwarzschild, Z.~C. Lipton, and J.~Z. Kolter, ``Tofu: A task of fictitious unlearning for llms,'' 2024.

\bibitem{cha2024learning}
S.~Cha, S.~Cho, D.~Hwang, H.~Lee, T.~Moon, and M.~Lee, ``Learning to unlearn: Instance-wise unlearning for pre-trained classifiers,'' 2024.

\bibitem{Chaudhuri2011}
K.~Chaudhuri, C.~Monteleoni, and A.~D. Sarwate, ``Differentially private empirical risk minimization,'' \textit{Journal of Machine Learning Research}, vol.~12, pp. 1069--1109, 2011.

\bibitem{Dwork2006Differential}
C.~Dwork, ``Differential privacy,'' in \textit{International Colloquium on Automata, Languages, and Programming}, Springer, 2006, pp. 1--12.

\bibitem{Dwork2014TheAF}
C.~Dwork and A.~Roth, ``The algorithmic foundations of differential privacy,'' \textit{Found. Trends Theor. Comput. Sci.}, vol.~9, pp. 211--407, 2014. 

\end{thebibliography}
\end{document}